\documentclass[twoside]{article}
\usepackage[accepted]{aistats2022}
\usepackage[utf8]{inputenc} 
\usepackage[T1]{fontenc}    
\usepackage{comment}
\usepackage{hyperref}       
\usepackage{url}            
\usepackage{nicefrac}       
\usepackage{microtype}     
\newcommand{\algemph}[3]{\algcolor{#1}{#2}{#3}}

\usepackage{xcolor}
\usepackage{framed}
\colorlet{shadecolor}{pink} 
\usepackage{authblk}
\usepackage{adjustbox}
\usepackage{comment}

\usepackage{graphicx}
\usepackage{soul}
\usepackage{subcaption}
\usepackage{booktabs}
\usepackage{tablefootnote}

\usepackage{amsmath,amsthm,amssymb,amsfonts}
\usepackage{algorithm}
\usepackage{algorithmic}
\usepackage{enumerate}
\usepackage{cleveref}
\usepackage{comment}
\usepackage{bm}
\usepackage{pifont}

\theoremstyle{plain}
\newtheorem{theorem}{Theorem}[]

\newtheorem{lemma}[]{Lemma}

\newcommand{\vertiii}[1]{{\left\vert\kern-0.25ex\left\vert\kern-0.25ex\left\vert #1 
\right\vert\kern-0.25ex\right\vert\kern-0.25ex\right\vert}}

\newtheorem{assumption}{Assumption}[]

\usepackage{algorithm}
\usepackage[ruled,algo2e]{algorithm2e}
\usepackage{hyperref}
\hypersetup{colorlinks=true,linkcolor=blue,filecolor=magenta,urlcolor=cyan,}

\definecolor{antiquewhite}{rgb}{0.98, 0.92, 0.84} 
\definecolor{blizzardblue}{rgb}{0.67, 0.9, 0.93}
\newcommand{\algcolor}[3]{\hspace*{-\fboxsep}\colorbox{#1}{\parbox{#2\linewidth}{#3}}}

\begin{document} 

\twocolumn[

\aistatstitle{Local SGD Optimizes Overparameterized Neural Networks in Polynomial Time}

\aistatsauthor{ Yuyang Deng \And Mohammad Mahdi Kamani\And  Mehrdad Mahdavi}

\aistatsaddress{ The Pennsylvania State University \And  Wyze Labs Inc. \And The Pennsylvania State University } ]

\begin{abstract}
In this paper we prove that Local (S)GD (or FedAvg) can optimize deep neural networks with Rectified Linear
Unit (ReLU) activation function in polynomial time. Despite the established convergence theory of Local SGD on optimizing general smooth functions in  communication-efficient distributed optimization, its  convergence on non-smooth ReLU networks  still eludes full theoretical understanding. The key property used in many Local SGD analysis on smooth function is gradient Lipschitzness, so that the gradient on local models will not drift far away from that on averaged model. However, this decent property does not hold in networks with non-smooth ReLU activation function.  We show that, even though ReLU network does not admit gradient Lipschitzness property, the difference between  gradients on local models and average model will not change too much, under the dynamics of Local SGD. We  validate our theoretical results via extensive experiments. This work is the first to show the convergence of Local SGD on non-smooth functions, and will shed lights on the optimization theory of federated training of deep neural networks. 
\end{abstract}

\section{Introduction}
The proliferation of mobile devices and internet of things (IoT) have resulted in  immense growth of data generated by users, and offer huge potential in further advancement in ML if harnessed properly. However, due to regulations and concerns about data privacy,  collecting data from clients and training machine learning models on a central server is not plausible. To decouple the ability to do machine learning without directly accessing  private data of users,  the Local SGD (a.k.a. Federated Averaging (FedAvg)) algorithm proposed in~\cite{mcmahan2017communication}  to train deep neural networks in a communication efficient manner, without leaking users' data. In Local SGD, the goal is to minimize a finite sum problem under the orchestration of a central server, where each component function is the empirical loss  evaluated on each client's local data. Local clients  perform SGD on their own local models  and after every $\tau$ steps, server  synchronizes the models by aggregating locally updated models and averaging them. This simple idea has been shown to be effective in reducing the number of communication rounds, while enjoying the same convergence rate as fully synchronous counterpart, and become the key optimization method in many federated learning  scenarios. We refer readers to several recent surveys~\cite{konevcny2016afederated,konevcny2016bfederated,li2020federated,mcmahan2021advances} and the references
therein for a non-exhaustive list of the research.  

Although  significant advances have been made on  understanding  the convergence theory of Local SGD~\cite{stich2018local,khaled2020tighter,haddadpour2019convergence,haddadpour2019local,li2019convergence,woodworth2020local,woodworth2020minibatch},  however, these works mostly focus on general smooth functions. It has been observed that Local SGD can also efficiently optimize specific family of  non-smooth functions, e.g., deep ReLU networks~\cite{mcmahan2017communication,zhao2018federated,li2018federated,hard2018federated}. Up until now, the theoretical understanding of  Local SGD  on optimizing this class  of  non-smooth functions remains elusive. Inspired by this, we focus on  rigorously understanding the  convergence of Local GD or Local SGD when utilized to  optimize non-smooth objectives.

%Then, a significant question remains unsolved: why can local SGD efficiently optimize a class of non-smooth function: ReLU neural network?  

%we optimize the virtual averaged model using local gradients and hence 

While numerous studies investigated  the behavior of single machine SGD on optimizing deep neural networks~\cite{du2018gradient,du2019gradient,allen2018learning,allen2019convergence,arora2019fine,zou2019improved,zou2020gradient}, and established linear convergence when the neural network is wide enough, however, these results cannot be trivially generalized to Local SGD. In fact,  in  local methods, due to  local updating and periodic synchronization, the desired analysis should be more involved to bound the difference between local models and (virtual) averaged model. On general smooth functions, according to gradient Lipschitzness property, we know that local gradients are close to gradients on averaged model. However, due to non-smoothness of ReLU function, this idea is no longer applicable. This  naturally raises the question of understanding  
\textit{why Local (S)GD can optimize deep ReLU neural networks}, which we aim to answer in this paper.

\noindent\textbf{Contributions.} We show that, both Local GD and Local SGD can provably optimize \textbf{deep} ReLU networks with multiple layers  in polynomial time, under \textit{heterogeneous data allocation} setting, meaning that each  client has  training data sampled from a potentially different underlying distribution. In the deterministic setting, we prove that Local GD can optimize an $L$-layer ReLU network with $\Omega(n^{16} L^{12})$ neurons, with a linear convergence rate  ${O}(e^{-R})$, where $R$ is the total number of communication rounds. In the stochastic setting, we prove that Local SGD can optimize an $L$-layer ReLU network with $\Omega(n^{18} L^{12})$ neurons, with the rate of $O(e^{-R/R_0})$, where $R_0$ is some constant depending on the number of samples $n$ and neurons $m$.  To the best of our knowledge, this paper is the first to analyze the global convergence of the both Local GD and Local SGD methods on optimizing \textit{deep} neural networks with  ReLU activation, and the first to show that it can converge even on non-smooth functions.  To support our theory, we conduct  experiments on MNIST dataset and demonstrate that the results match with our theoretical findings.

From a technical perspective, a key challenge to establish the convergence of both methods appears to be the non-smoothness of objective. In fact, as mentioned before, in the analysis of Local SGD on general smooth functions, a crucial step is to leverage the gradient Lipschitzness property, such that we can bound the gap between gradients on local model and averaged model. However, deep ReLU networks do not admit such benign property which complicates bounding the drift between local models and virtual averaged model due to multiple local updates (i.e., infrequent synchronization). To overcome the difficulty resulting from non-smoothness, we discover a ``\textbf{semi gradient Lipschitzness}'' property that indicates  despite the non-smooth nature of ReLU function, its gradient still enjoys some almost-Lipschitzness geometry  and characterizes the second order Lipschitzness nature of the neural network loss. This allows us  to develop techniques to bound the local model deviation under the dynamics of Local (S)GD. 

\noindent\textbf{Notations.} We use boldface lower-case letters such as $\boldsymbol{x}$ and upper-case letters such as $\mathbf{W} $ to denote vectors and matrices, respectively. We use $\|\bm{v}\|$ to denote Euclidean norm of vector $\bm{v}$, and use $\|\mathbf{W} \|$ and $\|\mathbf{W} \|_{\mathrm{F}}$ to denote spectral and Frobenius norm of matrix $\mathbf{W} $, respectively. We use $\mathcal{N}(\mu,\delta)$ to denote the Gaussian distribution with mean $\mu$ and variance $\delta$. We also use $\mathbf{W} $ to denote the tuple of all $\mathbf{W} _1,...,\mathbf{W}_L$, i.e., $\mathbf{W}  = (\mathbf{W} _1,...,\mathbf{W}_L)$.    Finally, we use $\mathcal{B}(\mathbf{W} , \omega)$ to denote the Euclidean ball centered at $\mathbf{W}$ with radius $\omega$.
%Periodic synchronization in parallel SGD was initially investigated in~\cite{zhang2016parallel}.
\section{Related Work}
\noindent\textbf{Local SGD.}~Recently, the most popular idea to achieve communication efficiency in distributed/federated optimization is Local SGD or FedAvg, which is firstly proposed by McMahan et al~\cite{mcmahan2017communication} to alleviate communication bottleneck in the distributed machine learning via periodic synchronization, which is initially investigated empirically  in~\cite{zhang2016parallel} to improve  parallel SGD. Stich~\cite{stich2018local} gives the first proof that Local SGD can optimize smooth strongly convex function at the rate of $O\left(\frac{1}{KT}\right)$, with only $O(\sqrt{KT})$ communication rounds. \cite{khaled2020tighter} refine the Stich's bound, which reduces the $O(\sqrt{T})$ communication rounds to $\Omega(K)$. Haddadpour et al~\cite{haddadpour2019local} give the first analysis on the nonconvex (PL condition) function, and proposed an adaptive synchronization scheme. Haddadpour and Mahdavi~\cite{haddadpour2019convergence} gave the analysis of Local GD and SGD on smooth nonconvex functions, under non-IID data allocation. Li et al~\cite{li2019convergence}  also prove the convergence of FedAvg on smooth strongly convex function under non-IID data setting . \cite{woodworth2020local,woodworth2020minibatch} do the comparison between mini-batch SGD and Local SGD by deriving the lower bound for mini-batch SGD and Local SGD, in both homogeneous and heterogeneous data settings. For some variant algorithm, Karimireddy et al~\cite{karimireddy2019scaffold} propose SCAFFOLD algorithm which mitigates the local model drifting and hence speed up the convergence. Yuan and Ma~\cite{yuan2020federated} borrow the idea from acceleration in stochastic optimization, and  propose the first accelerated federated SGD, which further reduced the communication rounds to  $O(K^{1/3})$. 

\noindent\textbf{Convergence Theory of Neural Network.}~ The empirical success of (deep) neural networks motivated the researchers  to study the theoretical foundation behind them. Numerous studies take efforts to establish the convergence theory of overparameterized neural networks. While earlier  works study the simple two layer network as the starting point~\cite{tian2017analytical,du2018power,li2017convergence,zhong2017recovery,brutzkus2017globally}, but these papers make strong assumption on input data or sophisticated initialization strategy. Li and Liang~\cite{li2018learning} study the two layer network with cross-entropy loss, and for the first time show that if the network is overparameterized enough, SGD can find the global minima in polynomial time. Furthermore, if the input data is well structured, the guarantee for generalization can also be achieved. Du et al~\cite{du2018gradient} derive the global linear convergence of two-layer ReLU network with $l_2$ regression loss. They also extend their results to deep neural network in~\cite{du2019gradient}, but they assume the activation is smooth. Allen-Zhu et al~\cite{allen2019convergence} firstly prove the global linear convergence of deep RelU network, and derive a key semi-smoothness property of ReLU DNN, which advances the analysis tool for ReLU network. Zou and Gu~\cite{zou2019improved} further improve Allen-Zhu's result. They reduce the width of the network to a small dependency on the number of training samples, by deriving a tighter gradient upper bound. Recently, some works further reduce this dependency to cubic, quadratic and even linear~\cite{noy2021convergence,nguyen2020global,nguyen2021proof}. 

\noindent\textbf{Local (S)GD on Neural Network.}~Recently, Huang et al~\cite{huang2021fl} study the convergence of Local GD on 2-layer ReLU network, which is the most relevant work to ours. However, besides the analysis methods which are significantly different,~\cite{huang2021fl} only considers deterministic algorithm (Local GD) on a simple two-layer network. In this paper, we establish convergence for both Local GD and Local SGD  on an $L$-layer deep ReLU network. 

\section{Problem Setup}

We consider a distributed setting with $K$ machines. Let $S_i = \{(\bm{x}_1^i,y_1^i),...,(\bm{x}_n^i,y_n^i)\}$ denote the set of all $n$ training data allocated at client $i$ . We further let $S = \bigcup_{i=1}^K S_i$ to be the union of all clients' data. The goal is to solve the following finite sum minimization problem in a distributed manner:
\begin{equation*}
        \min_{\mathbf{W}}  L(\mathbf{W}) = \frac{1}{K}\sum_{i=1}^K  L_i(\mathbf{W}),
\end{equation*}
where $ L_i(\mathbf{W} ) = \frac{1}{n}\sum_{(\bm{x},y)\in S_i}\ell (\mathbf{W} ;\bm{x},y)$ is the loss function evaluated on $i$th client data based on loss function $\ell()$. The description of the network architecture and loss function type are presented next.

\noindent\textbf{Deep ReLU network.} We consider a $L$-layer neural network architecture with ReLU activation function:
\begin{align}
    f(\mathbf{W} ,\mathbf{V}, \bm{x}) = \mathbf{V} \sigma (\mathbf{W}_L \sigma(\mathbf{W}_{L-1}\cdots \sigma(\mathbf{W}_1 \bm{x})))\nonumber
\end{align}
where $\sigma(x) = \max(x,0)$, $\mathbf{W}_l \in \mathbb{R}^{m\times m}$ is the weight matrix of $l$th layer (we set $\mathbf{W}_1 \in \mathbb{R}^{m\times d}$), $\bm{x}\in \mathbb{R}^{d}$ is the input data.  For ease of exposition, we assume the number of neurons is same for all layers. Also, following the prior
studies~\cite{du2018gradient,allen2019convergence,zou2019improved}, we fix the top layer $\mathbf{V} $, and only train the parameters of the hidden layers $\mathbf{W} = (\mathbf{W}_1, \mathbf{W}_2, \ldots, \mathbf{W}_L)$.

We consider regression setting with squared losss
$     \ell (\mathbf{W},\mathbf{V} ;\bm{x},\bm{y}) = \frac{1}{2} \|f(\mathbf{W} ,\mathbf{V} ,\bm{x})-\bm{y}\|^2\nonumber
$ where  the gradient of $L_i(\mathbf{W} )$ w.r.t. $\mathbf{W}_l$ can be derived as:
\begin{equation*}
    \begin{aligned}
       &\nabla_{\mathbf{W}_{l}}L_i(\mathbf{W} )  =  \frac{1}{n} \sum_{(\bm{x}_j,\bm{y}_j) \in S_i} \mathbf{D}_{j,l} \mathbf{B}_{j,l+1}^\top (f_j - \bm{y}_j) f_{j,l-1},
\end{aligned}
\end{equation*}
where 
\begin{equation*}
    \begin{aligned}
    f_{j,l} &=   \sigma (\mathbf{W}_l \sigma(\mathbf{W}_{l-1}\cdots \sigma(\mathbf{W}_1 \bm{x}_j))),\\ 
    f_j  &= \mathbf{V} \sigma (\mathbf{W}_L \sigma(\mathbf{W}_{L-1}\cdots \sigma(\mathbf{W}_1 \bm{x}_j))),\\ \mathbf{B}_{j,l+1} &= \mathbf{V}\mathbf{D}_{j,L}\mathbf{W}_L\cdots\mathbf{D}_{j,l+1}\mathbf{W}_{l+1}
    \end{aligned}
\end{equation*}
and $\mathbf{D}_{j,l} \in \mathbb{R}^{m\times m}$ is a diagonal matrix with entries $\mathbf{D}_{j,l}(r,r) = \mathbf{1}[(\mathbf{W}_l f_{j,l-1})_r\geq 0]$ for $r \in [m]$. For ease of exposition, we will express  $\nabla_{\mathbf{W}}L_i(\mathbf{W} )$  as the following tuple:
\begin{equation*}
     \nabla_{\mathbf{W} }L_i(\mathbf{W} ) = ( \nabla_{\mathbf{W}_{1}}L_i(\mathbf{W} ),\cdots,\nabla_{\mathbf{W}_{L}}L_i(\mathbf{W} ) ) \nonumber.
\end{equation*}

\paragraph{Algorithm description.}To mitigate the communication bottleneck in distributed optimization, a popular idea is to update models locally via GD or SGD, and then average them periodically~\cite{mcmahan2017communication,stich2018local}. The Local (S)GD algorithm proceeds for $T$ iterations, and at $t$th iteration, the $i$th client locally performs the GD or SGD on its own model $\mathbf{W}^{(i)}(t)$:

{\small\begin{align}
       &\text{Local GD:} \ { \mathbf{W} }^{(i)}(t+1) = {\mathbf{W} }^{(i)}(t) - \eta \nabla_{\mathbf{W} }      L_i\left({\mathbf{W} }^{(i)}(t)\right),   \nonumber \\
     & \text{Local SGD:} \  { \mathbf{W} }^{(i)}(t+1) = {\mathbf{W} }^{(i)}(t) - \eta  \bm{G}^{(t)}_i ,\nonumber
\end{align}}
where $\bm{G}^{(t)}_i$ is the stochastic gradient such that $\mathbb{E}[\bm{G}^{(t)}_i] = \nabla_{\mathbf{W} } L_i\left({\mathbf{W} }^{(i)}(t)\right)$. After $\tau$ local updates (i.e., $t$ divides $\tau$), the server aggregates local models $\mathbf{W} ^{(i)}(t+1), i=1, \ldots, K$ and performs the next global model according to:
\begin{align*}
    \mathbf{W} (t+1) = \frac{1}{K} \sum_{i=1}^K \mathbf{W} ^{(i)}(t+1).
\end{align*}
Then, the server sends the averaged model back to  local clients, to update their local models and  the procedure is repeated for $T/\tau$ stages. This idea can significantly reduce the communications rounds by a factor of $\tau$, compared to fully synchronized GD/SGD. Even though it is a simple algorithm, and has been employed for distributed neural network training for a long time, we are not aware of any prior theoretical work that analyzes its convergence performance on deep ReLU neural networks.  We note that  the aggregated model at server cannot be treated as $\tau$ iterations of synchronous SGD, since each local update contains a bias with respect to the global model which necessities to  bound the drift among local and global models. The bias issue becomes even more challenging when non-smooth ReLU is utilized compared to the existing studies that focus on smooth objectives.

\begin{algorithm2e}[t]
	\renewcommand{\algorithmicrequire}{\textbf{Input:}}
	\DontPrintSemicolon
    \caption{\texttt{Local (S)GD}}
	\label{algorithm: GD}
 
    	\textbf{Input:} Synchronization gap $\tau$ , Number of iterations $T$.  Initialization network parameter $\bm{W}(0) \sim \mathcal{N}(0,2/m\bm{I})$ and $\bm{v} \sim \mathcal{N}(0,  \bm{I}/d)$
	\\
 
	\textbf{parallel} \For{$i = 1,...,K$}{
	 
	 \For{$t = 1,...,T$}{
  
 	   \algemph{antiquewhite}{0.85}{       ${\bm{W}}^{(i)}(t+1) = {\bm{W}}^{(i)}(t) - \eta \nabla_{\bm{W}}      L_i\left({\bm{W}}^{(i)}(t)\right) $ \qquad  \# Local GD  }\\   
 	   \algemph{blizzardblue}{0.85}{
 	   Sample a data $(\Tilde{\bm{x}}, \Tilde{y})$ uniformly from $S_i$.\\
 	    Compute $\bm{G}_i^{(t)} = n \nabla_{\bm{W}}      \ell\left({\bm{W}}^{(i)}(t);\Tilde{\bm{x}}, \Tilde{y} \right)$.\\
 	 ${\bm{W}}^{(i)}(t+1) = {\bm{W}}^{(i)}(t) - \eta \bm{G}_i^{(t)} $  \qquad \qquad \qquad \quad \# Local SGD  \\   }
	 \If{$t$ divides $\tau$}{
	        all nodes send their local parameter ${\bm{W}}^{(i)}(t+1)$ to server.\\
           $ \bm{W}(t+1) = \frac{1}{K}\sum_{i=1}^K {\bm{W}}^{(i)}(t+1)$;\\
 	         
 	        send $ \bm{W}(t+1)$ to all nodes to update their local models. \\
 	        each client initializes its local models: $\bm{W}^{(i)}(t+1)= \bm{W}(t+1)$.} 
	 } 
	
  }

\end{algorithm2e}

% \begin{algorithm2e}[t]
% 	\renewcommand{\algorithmicrequire}{\textbf{Input:}}
% 	\DontPrintSemicolon
%     \caption{\texttt{Local SGD}}
% 	\label{algorithm: SGD}
 
%     	\textbf{Input:} Synchronization gap $\tau$, Number of iterations $T$. Initialization network parameter $\bm{W}(0) \sim \mathcal{N}(0,2/m\bm{I})$ and $\bm{v} \sim \mathcal{N}(0,  \bm{I}/d)$
% 	\\
 
% 	\textbf{parallel} \For{$i = 1,...,K$}{
	 
% 	 \For{$t = 1,...,T$}{
  
%  	 {Sample a data $(\Tilde{\bm{x}}, \Tilde{y})$ uniformly from $S_i$.\\
%  	 Compute $\bm{G}_i^{(t)} = n \nabla_{\bm{W}}      \ell\left({\bm{W}}^{(i)}(t);\Tilde{\bm{x}}, \Tilde{y} \right)$.\\
%  	 ${\bm{W}}^{(i)}(t+1) = {\bm{W}}^{(i)}(t) - \eta \bm{G}_i^{(t)} $   \\   }
% 	 \If{$t$ divides $\tau$}{
% 	        all nodes send their local parameter ${\bm{W}}^{(i)}(t+1)$ to server.\\
%           $ \bm{W}(t+1) = \frac{1}{K}\sum_{i=1}^K {\bm{W}}^{(i)}(t+1)$;\\
 	         
%  	        send $ \bm{W}(t+1)$ to all nodes to update their local models. \\
%  	        each client initializes its local models: $\bm{W}^{(i)}(t+1)= \bm{W}(t+1)$.} 
% 	 } 
	
%   }

% \end{algorithm2e} 

\section{Main Results}\label{sec: convergence}
In this section, we present the convergence rates. We start with making the following standard separability assumption~\cite{allen2019convergence,zou2019improved} on the training data.
\begin{assumption}\label{assumption: data separation}
For any $\bm{x}\in S, \|\bm{x}\| = 1$, and for any $\bm{x}, \bm{x}' \in S$, $\|\bm{x}-\bm{x}'\|\geq \phi$.
\end{assumption}

The following theorem establishes the convergence rate of Local GD on deep ReLU network:
\begin{theorem}[Local GD]\label{theorem: local GD}
For Local GD, under Assumption~\ref{assumption: data separation}, if we choose $m \geq \frac{K d n^{16} L^{12}(\log m)^3}{\phi^5}$, $\eta = O\left(\frac{d n^2}{m\phi \tau}\right)$   
 then with probability at least $1-e^{-\Omega( (\log m)^2)}$ it holds that
\begin{align*}
   & L(\bm{W}(T)) \leq  e^{-\Omega(R)} L(\bm{W}(0)) , 
\end{align*}  
where $R = \frac{T}{\tau}$ is the total number of communication rounds, and $\bm{W}(T) = \frac{1}{K}\sum_{i=1}^K \bm{W}^{(i)}(T)$.
\end{theorem}
The proof of Theorem~\ref{theorem: local GD} is provided in Appendix~\ref{app: GD}. As expected, the fastest convergence rate is attained
when the  synchronization gap $\tau$ is one. Theorem~\ref{theorem: local GD} however precisely  characterizes how large the number of neurons needs to picked to guarantee linear convergence rate. Here we require the width of network $m$ to be $O(K n^{16}L^{12})$ to achieve linear rate in terms of communication rounds, which is linear in the number of clients and polynomial in $n$ and $L$. The most relevant work to this paper is Huang et al~\cite{huang2021fl}, where they consider two-layer ReLU network, and achieve and $O(e^{-R/K})$ convergence rate with $\Omega(n^4)$ neurons.  Their convergence rate is strictly worse than us, while they require smaller number of neurons because they only consider simple two-layer architecture. An interesting observation from above rate is that the number of neurons per layer is  polynomial in the  number of layers which is also observed in our empirical studies. This implies that by adding to the depth of model, we also need to increase the number of neurons at each layer accordingly. We note that compared to analysis of single machine GD on deep ReLU networks~\cite{zou2019improved,nguyen2021proof}, the width obtained here is worse, and we leave the improvement on either the dependency on $n$ or $K$ as a future work. 

% \begin{remark}
% Notice that the obtained rates hold under a condition for $\tau$: $ \tau \leq \frac{T^{(1-{\tau}/{T})}}{ m (\log T)^2}$. Since the RHS of the inequality increases as $T$ increases, then the condition can hold for sufficiently large $T$. If we fix $T$, then $\tau$ is inversely proportional to $m$. This observation suggests that, if the network is complex (i.e., $m$ is large), then Local (S)GD needs to communicate more frequently (i.e., $\tau$ is small) to guarantee the convergence. 
% \end{remark}

% An immediate implication of Theorem~\ref{theorem: local GD} is:
% \begin{corollary}\label{coro: GD}
% Continuing with Theorem~\ref{theorem: local GD}, if we choose $\tau = \sqrt{T/K}$, and $T$ is large enough such that $T^{1-\sqrt{T/K}} \geq m (\log T)^2 \sqrt{T/K}$, which satisfies the condition in Theorem~\ref{theorem: local GD}, then we recover the rate:
% {\small\begin{align*}
%   & L(\bm{W}(T)) \leq  \tilde{O}\left( \frac{\exp\left( \frac{1}{\phi n^{9/2} (\log m)^{3/2}}\right) m^{1/2} n^4 }{\phi^{5/2} d^{1/2}  } \frac{L(\bm{W}(0))}{ \sqrt{KT}}\right).
% \end{align*} }
% \end{corollary}
% By properly choosing synchronization gap $\tau$, we achieve linear speedup in the number of working clients.

Now we proceed to establish the convergence rate  of Local SGD:

\begin{theorem}[Local SGD]\label{theorem: local SGD}
For Local SGD, under Assumption~\ref{assumption: data separation}, if we choose $m \geq \frac{K d n^{18} L^{12} (\log m)^5}{\phi^3}$, $\eta = O\left( \frac{d \phi}{m\tau n^3 \log^2 m}\right)$   
 then with probability at least $1-e^{-\Omega( (\log m)^2)}$ it holds that
\begin{align*}
   & L(\bm{W}(T)) \leq (n\log^2m) \cdot e^{-\Omega(R/R_0)} L(\bm{W}(0)) , 
\end{align*}  
where $R = \frac{T}{\tau}$ is the total number of communication rounds, $R_0 = \frac{ n^5\log^2m}{\phi^2}$, and $\bm{W}(T) = \frac{1}{K}\sum_{i=1}^K \bm{W}^{(i)}(T)$.
\end{theorem}

% The proof of Theorem~\ref{theorem: local SGD} is provided in Appendix~\ref{app: SGD}.
% An immediate implication of Theorem~\ref{theorem: local SGD} is the following.
% \begin{corollary}\label{coro: SGD}
% Continuing with Theorem~\ref{theorem: local SGD}, if we choose $\tau = \sqrt{T/K}$, and $T$ is large enough such that $T^{1-\sqrt{T/K}} \geq m (\log T)^2 \sqrt{T/K}$ which satisfies the condition in Theorem~\ref{theorem: local SGD}, then we recover the rate:
% {\small\begin{align*}
%     L(\bm{W}(T)) \leq  \tilde{O}\left( \frac{\exp\left(  \frac{1 }{ n^{5/2}\phi^{1/2} (\log m)^{3/2}  }\right)d^{1/2}m^{1/2}n^{11 }   }{\phi^{5/2} } \frac{L(\bm{W}(0))}{ \sqrt{KT}}\right).
% \end{align*} }
% \end{corollary}

\noindent\textbf{Comparison to related bounds on Local SGD.} ~\cite{haddadpour2019convergence}~established an $O(1/\sqrt{KT})$ rate with $O(\sqrt{KT})$ communication rounds on general smooth nonconvex functions, while our result enjoys faster rate and better communication efficiency. We would also like to emphasize that our setting is more difficult, since 1) we study nonconvex and non-smooth functions; 2) we prove a global convergence, but their result only guarantees the convergence to a first order stationary point, and 3) our result is stated for last iterate, but theirs only guarantees that at least one of the history iterates vissits local minima.
  
 \noindent\textbf{Comparison to related work on ReLU networks.} Since we are not aware of any related work of Local SGD on ReLU network, here we only discuss single machine algorithms.  Compared to single machine SGD on optimizing ReLU network, the most analogous work to ours is~\cite{zou2019improved}, since both our and their analysis adapt the proof framework from~\cite{allen2019convergence}. They achieve linear convergence with $\Omega(n^{17})$ neurons for achieving linear convergence, while we need $\Omega(n^{18})$ neurons. We also noticed that recent works~\cite{nguyen2020global,nguyen2021proof} have reduce the network width to a significantly small number, so we leave improving our results by adapting a finer analysis as future work.
\section{Overview of Proof Techniques}
 In this section we will present an overview of our proof strategy for deterministic setting (Local GD). The stochastic setting shares the similar strategy. We let $\mathbf{W} (t) = \frac{1}{K}\sum_{i=1}^K \mathbf{W}^{(i)}(t)$ denote the virtual averaged iterates. We use $t_c$ to denote the latest communication round, also the $c$th communication round.

\subsection{Main Technique}
Our proof involves three main ingredients, namely (i) \textbf{semi gradient Lipschitzness}, (ii) \textbf{shrinkage of local loss}, and (iii) \textbf{local model deviation analysis} as we discuss briefly below.

\noindent\textbf{Semi Gradient Lipschitzness.}~\footnote{Notice that this is not the semi-smoothness property derived by Allen Zhu et al~\cite{allen2019convergence}, even though we also need that property in analysis.}In the analysis of Local SGD on general smooth functions, one key step is to utilize the gradient Lipschitzness property, such that we can bound the gap between gradients on local model and averaged model by: $\left\|\nabla L(\mathbf{W}) - \nabla L(\tilde{\mathbf{W}}) \right\| \leq H\left\| \mathbf{W} -  \tilde{\mathbf{W}}   \right\|$. However, ReLU network does not admit such benign property. Alternatively, we discover a ``semi-gradient Lipschitzness'' property. For any parameterization $\mathbf{W} $ and $\tilde{\mathbf{W}} $ such that ${\mathbf{W}},\mathbf{W} \in \mathcal{B}(\mathbf{W}(0),\omega)$:

{\small{
\begin{equation*}
\begin{aligned}
    &\frac{1}{K}\sum_{i=1}^K \left\| \nabla_{\mathbf{W}}L_i(\mathbf{W})- \nabla_{\tilde{\mathbf{W}} }L_i  (\tilde{\mathbf{W}})\right\|^2_{\mathrm{F}} \\
    &\leq O\left( \frac{m L^4}{d} \left\|\mathbf{W} - \tilde{\mathbf{W}}\right\|_2^2\right)+O\left(\frac{\omega^{2/3} L^5 {m\log m }}{d}\right) L(\tilde{\mathbf{W}}),
\end{aligned}
\end{equation*}
}}

This inequality demonstrates that for any two models lying in the small local perturbed region of initialization model, ReLU network almost achieves gradient Lipschitzness, up to some small additive zeroth order offset. That is, if we can carefully move local models $\mathbf{W} ^{(i)}(t)$ such that they do not drift from the initialization and virtual average model $\mathbf{W}(t)$ too much,  then the gradient at local iterate $\nabla_{\mathbf{W} ^{(i)}}L_i(\mathbf{W} ^{(i)}(t))$ is guaranteed to be close to the gradient at virtual averaged iterate $ \nabla_{\mathbf{W} }L_i  (\mathbf{W} (t))$.

\noindent\textbf{Shrinkage of Local Loss.} Another key property of local loss is that the local loss is strictly decreasing, compared to the latest communication round. We show that with high probability, if we properly choose learning rate, the following inequality holds:
for Local GD:
  \begin{align*}
     L_i(\mathbf{W}^{(i)}(t)) \leq  L_i(\mathbf{W}^{(i)}(t-1)) \leq \cdots \leq L_i(\mathbf{W}^{(i)}(t_c)),
 \end{align*} 
and for Local SGD:
\begin{align*}
   L_i(\mathbf{W}^{(i)}(t)) \leq  \exp \left(\frac{ \phi}{m  n^{2.5} \log^2 m  } \right)  L_i( \mathbf{W}(t_c)),
\end{align*}
where $t_c \leq t\leq t_c + \tau -1$, and $t_c$ is the latest communication round of $t$. This nice property will enable us to reduce the loss at any iteration to its latest communication round.

\noindent\textbf{Local Model Deviation Analysis.} During the dynamic of Local (S)GD, the local models will drift from the virtual averaged model, so the other key technique in Local (S)GD analysis is to bound local model deviation $\|\mathbf{W} ^{(i)}(t)  -\mathbf{W} (t) \|_{\mathrm{F}}$. However, in the highly-nonsmooth ReLU network, this quantity is not a viable error to control. Hence, inspired by~\cite{huang2021fl}, we consider the deviation $\|\mathbf{W} ^{(i)}(t)  -\mathbf{W} (t_c) \|_{\mathrm{F}}$, where $t_c$ is the latest communication round of $t$, and derive the deviation bound as:\vspace{-0.5cm}

{\small\begin{align*}
  &\frac{1}{K}\sum_{i=1}^K \left\| \mathbf{W} ^{(i)}(t) -  \mathbf{W} (t_c) \right\|^2_{\mathrm{F}} \leq O\left( \frac{\eta^2 \tau^2mn}{d}\right) L(\mathbf{W}(t_c)).
\end{align*}}
\vspace{-0.5cm}

Here we bound the local model deviation by the loss at the last communication round, which is a key step that enables us to achieve linear rate.

\subsection{Sketch of the Proof}
In this section we are going to present the overview of our key proof techniques. The detailed proofs are deferred to appendix.  Before that,
we first mention two lemmas that facilitate our analysis.
%The proof framework is similar to Allen-Zhu et al~\cite{allen2019convergence} and Zou and Gu~\cite{zou2019improved}, which relies on the following two lemmas: 
\begin{lemma}[Semi-smoothness \cite{allen2019convergence}] \label{lemma:semi_smooth}
Let 
\begin{align*}
\omega \in \big[\Omega\big(1/(d^{3/2}m^{3/2} \log^{3/2}(m))\big),O\big(1/( \log^{3/2}(m))\big)\big].
\end{align*}
Then for any two weights $\hat{\mathbf{W} } $ and $\tilde {\mathbf{W} } $ satisfying $\hat{\mathbf{W} },\tilde{\mathbf{W} }\in \mathcal{B}({\mathbf{W} }^{(0)},\omega)$, with probability at least $1-\exp(-\Omega(m\omega^{3/2}L))$, there exist two constants $C'$ and $C''$ such that
\begin{align}\label{eq:semi_smooth}
L(\tilde{\mathbf{W} })&\le L(\hat{\mathbf{W} }) + \langle \nabla L(\hat {\mathbf{W} }),\tilde {\mathbf{W} } - \hat {\mathbf{W} }\rangle  \\
& \quad + C'\sqrt{L(\hat {\mathbf{W} })}\cdot\frac{\omega^{1/3} \sqrt{m\log(m)}}{\sqrt{d}}\cdot \|\tilde {\mathbf{W} } - \hat {\mathbf{W} }\| \\
& \quad+ \frac{C''m}{d}\|\tilde {\mathbf{W} } - \hat {\mathbf{W} }\|^2.
\end{align} 
\end{lemma}

 \begin{lemma}[Gradient bound~\cite{zou2019improved}]\label{lemma:grad bounds}
Let $\omega = O\big(\phi^{3/2}n^{-3}L^{-6} \log^{-3/2}(m)\big)$, then for all $\mathbf{W} \in\mathcal{B}(\mathbf{W}{(0)},\omega)$, with probability at least $1-\exp\big(-\Omega(m\phi/(dn)))$, it holds that
\begin{align*}
\|\nabla &L(\mathbf{W} )\|_{\mathrm{F}}^2\le O\big(m  L(\mathbf{W} )/d\big),\\
\|\nabla &L(\mathbf{W} )\|_{\mathrm{F}}^2\ge \Omega \big(m\phi L(\mathbf{W} )/(dn^2)\big).
\end{align*} 
\end{lemma}
The above two lemmas demonstrate that, if the network parameters lie in the  ball centered at initial solution with radius $\omega$, then the network admits local smoothness, and there is no critical point in this region.

The whole idea of the proof is that, we firstly assume each local iterates and virtual averaged iterates lie in the $\omega$-ball centered at initial model, so that we can apply the benign properties (semi smoothness, bounded gradients and semi gradient Lipschitzness) of objective function. Then, with these nice properties we are able to establish the linear convergence of the objective as claimed. Lastly, we verify the correctness of bounded local iterates and virtual averaged iterates assumption. 

The proof is conducted via induction. The inductive hypothesis is as follows: for any $h\leq t$, we assume the following statements holds for $\omega = O\big(\phi^{3/2} n^{-6} L^{-6} \log^{-3/2}(m)\big)$:
 \begin{align}
      \textbf{(I)} &\quad \left \| \mathbf{W}(h) -  \mathbf{W}(0)\right\| \leq \omega, \nonumber\\
      & \quad \left \| \mathbf{W}^{(i)}(h) -  \mathbf{W}(0)\right\| \leq \omega, \quad \forall i \in [K]\nonumber .\\ 
      \textbf{(II)} &\quad L(\mathbf{W}(t_c))   \le\left(1-  \Omega\left(\frac{\eta \tau m \phi}{dn^2}\right)\right)^{c} L( \mathbf{W}(0))\nonumber.
 \end{align}
 The first statement indicates that the virtual iterates do not drift too much from the initialization, under Local GD's dynamic, if we properly choose learning rate and synchronization gap. The second statement gives the linear convergence rate of objective value.  Now, we need to prove these two statements hold for $t+1$.

 \noindent\textbf{Step 1: Boundedness of virtual average iterates.} We first verify (\textbf{I}), the boundedness of virtual iterates during algorithm proceeding. The idea is to keep track of the dynamics of the average gradients on each local iterate. To do so, by the updating rule we have: 
\begin{align*}
      &\left \| \mathbf{W}(t+1) -  \mathbf{W}(0)\right\| \\
      &\leq \eta \sum_{j=1}^t\left \|\frac{1}{K}\sum_{i=1}^K \nabla L_i (\mathbf{W}^{(i)}(j))\right\| \nonumber\\
      &\leq \eta\sum_{j=1}^t \frac{1}{K}\sum_{i=1}^K O\left(\sqrt{\frac{m}{d}}\right)\sqrt{L_i(\mathbf{W}(j) )}    \nonumber \\ 
      & \leq \eta \tau \sum_{j=1}^c   O\left(\sqrt{\frac{m}{d}}\right)\sqrt{L(\mathbf{W}(t_c) )},
 \end{align*}
 where we apply the gradient upper bound (Lemma~\ref{lemma:grad bounds}) and the decreasing nature of local loss. 
 Now we plug in induction hypothesis \textbf{II} to bound $L(\mathbf{W}(t_c) )$:
    \begin{align*}
      &\left \| \mathbf{W}(t+1) -  \mathbf{W}(0)\right\|  \\
      & \leq \eta \tau \sum_{j=1}^c   O\left(\sqrt{\frac{m}{d}}\right)\sqrt{ \left(1-  \Omega\left(\frac{\eta \tau m \phi}{dn^2}\right)\right)^{c} L( \mathbf{W}(0))}\\
      & \leq \eta \tau  O\left(\sqrt{\frac{m}{d}}\right) \sum_{j=1}^c\left(1-  \Omega\left(\frac{\eta\tau  m \phi}{2dn^2}\right)\right)^{c}\sqrt{ L( \mathbf{W}(0))}\\
      & =     O\left(\frac{2\sqrt{d}n^2}{ \sqrt{m} \phi}\right) \sqrt{ L( \mathbf{W}(0))}.
 \end{align*}
Since we choose $m \geq \frac{K d n^{16} L^{12} \log^3 m}{\phi^5}$, it can be concluded that $ \left \| \mathbf{W}(t+1) -  \mathbf{W}(0)\right\|  \leq \omega$.

 \noindent\textbf{Step 2: Boundedness of local iterates.}  
 The next step is to show that local iterates are also lying in the local perturbed region of initial model. This can be done by tracking the dynamic of the gradients on individual local model:
  \begin{align*} 
      &\left \| \mathbf{W}^{(i)}(t+1) -  \mathbf{W}(0)\right\| \\
      &\leq \eta \sum_{j=1}^t\left \|  \nabla L_i (\mathbf{W}^{(i)}(j))\right\| \nonumber\\
      &\leq \eta \sum_{j=1}^t   O\left(\sqrt{\frac{m}{d}}\right)\sqrt{L_i(\mathbf{W}(j) )}    \nonumber \\ 
      & \leq \eta \tau \sum_{j=1}^c   O\left(\sqrt{\frac{m}{d}}\right)\sqrt{K L(\mathbf{W}(t_c) )}\\
      &\leq  O\left(\frac{2\sqrt{d}n^2}{ \sqrt{m} \phi}\right) \sqrt{K L( \mathbf{W}(0))} \leq \omega.
 \end{align*}

\noindent\textbf{Step 3: Linear convergence of objective value.}  We now switch to prove statement (\textbf{II}). Since we know that, $\|\mathbf{W} (t+1)-\mathbf{W} \|\leq \omega$, we can apply Lemma~\ref{lemma:semi_smooth}  by let $\tilde{\mathbf{W}} = \mathbf{W}(t_{c+1})$ and $\hat{\mathbf{W}} = \mathbf{W}(t_{c})$ and gradient bound (Lemma~\ref{lemma:grad bounds}). We have the following recursive relation over the loss at different communication stages: \vspace{-0.75cm}

{\small\begin{align*} 
&L(\mathbf{W}(t_{c+1})) \le \left(1-  \Omega\left(\frac{\eta \tau  m \phi}{dn^2}\right)\right) L(\mathbf{W}(t_c)) \\
& \quad + \frac{\eta }{2} \frac{1}{ K}\sum_{i=1}^K \sum_{t'=t_{c-1}}^{t_c-1}\left\| \nabla L_i( \mathbf{W}(t_c))-\nabla  L_i(\mathbf{W}^{(i)}(t')) \right\|^2_{\mathrm{F}}\\
&\quad+\eta C' \cdot\frac{\omega^{1/3}L^2\sqrt{m\log(m)}}{2\sqrt{d}}\\
& \quad\left(  \frac{1}{ K}\sum_{i=1}^K \sum_{t'=t_{c-1}}^{t_c-1}\left\|\nabla L( \mathbf{W}(t_c))-\nabla  L_i(\mathbf{W}^{(i)}(t'))  \right \|_{\mathrm{F}}^2\right),   
\end{align*}}

where $t_c$ is the latest communication round at iteration $t$. Now we can use semi gradient Lipschitzness property to reduce the difference between gradients to local model deviation. Further plugging in the local model deviation bound, and unrolling the recursion will  complete the proof:
\begin{align*} 
L(\mathbf{W}(t_{c}))  
&\le \left(1-  \Omega\left(\frac{\eta \tau  m \phi}{dn^2}\right)\right)L( \mathbf{W}(t_{c-1}))\\
&\leq e^{-\Omega(R)} L(\bm{W}(0))  . 
\end{align*}
as desired.

\section{Experiment}
In this section we present our experimental results to validate our theoretical findings. For this purpose, we run our experiments on MNIST dataset using a varying number of MLP layers with ReLU activation function on the hidden layer. We denote the number of neurons in the hidden layer with $m$. To run the experiments on a distributed setting, we create 50 clients.
% using the FedTorch\footnote{\url{https://github.com/MLOPTPSU/FedTorch}}.
Then, we distribute the MNIST dataset on these clients in IID (homogeneous) or non-IID (heterogeneous) ways. For IID setting, each client has training data i.i.d. sampled from the whole dataset.  For non-IID setting, we allocate only two classes of data to each client, and hence, different clients will have access to different distribution of data.

\begin{figure}[t]
    \centering
    \begin{subfigure}[b]{0.234\textwidth}
    \centering
    \includegraphics[width=\textwidth]{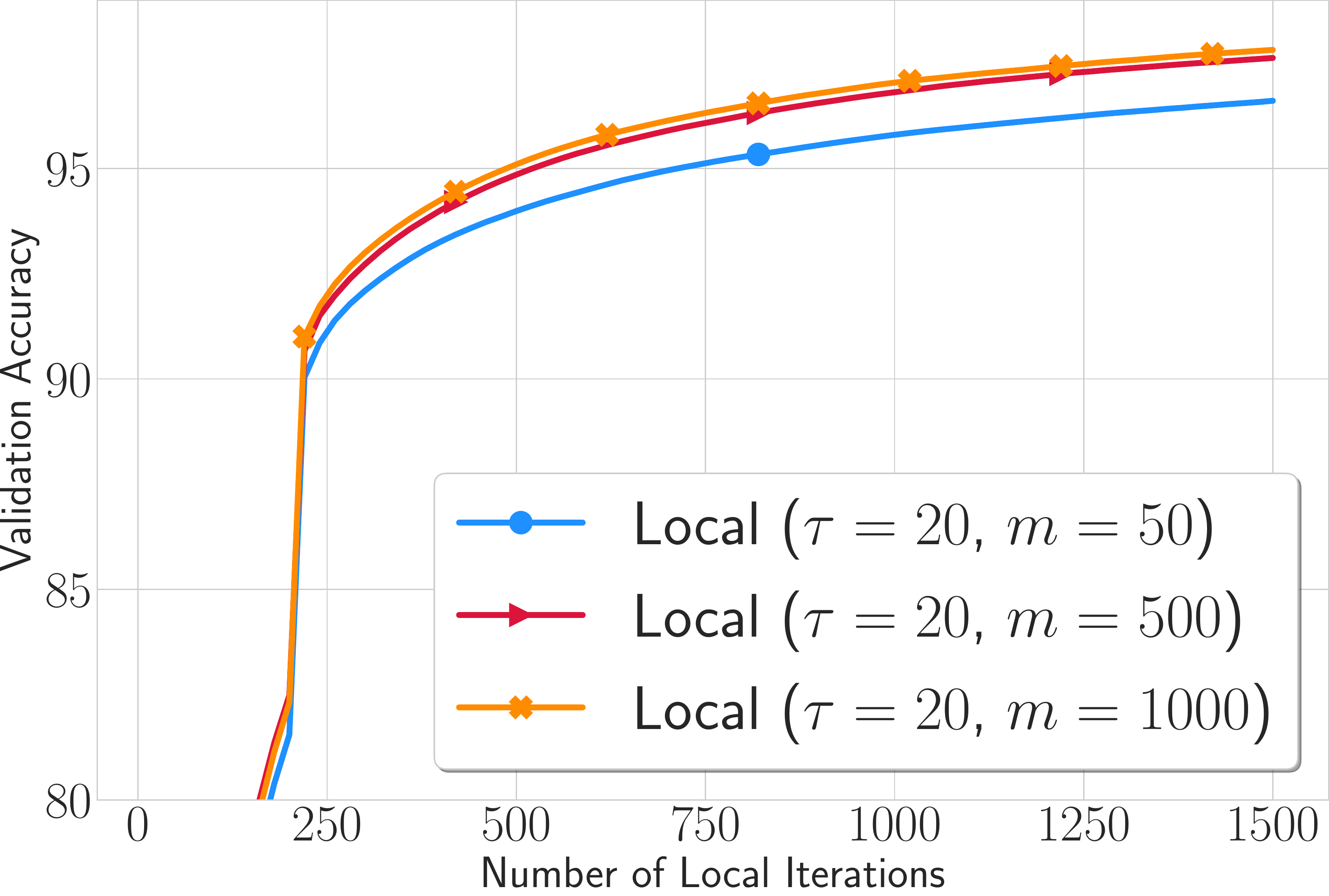}
    % \caption{Caption}
    \label{fig:my_label1}
    \end{subfigure}
    % \hspace{1cm}
    \begin{subfigure}[b]{0.234\textwidth}
    \centering
    \includegraphics[width=\textwidth]{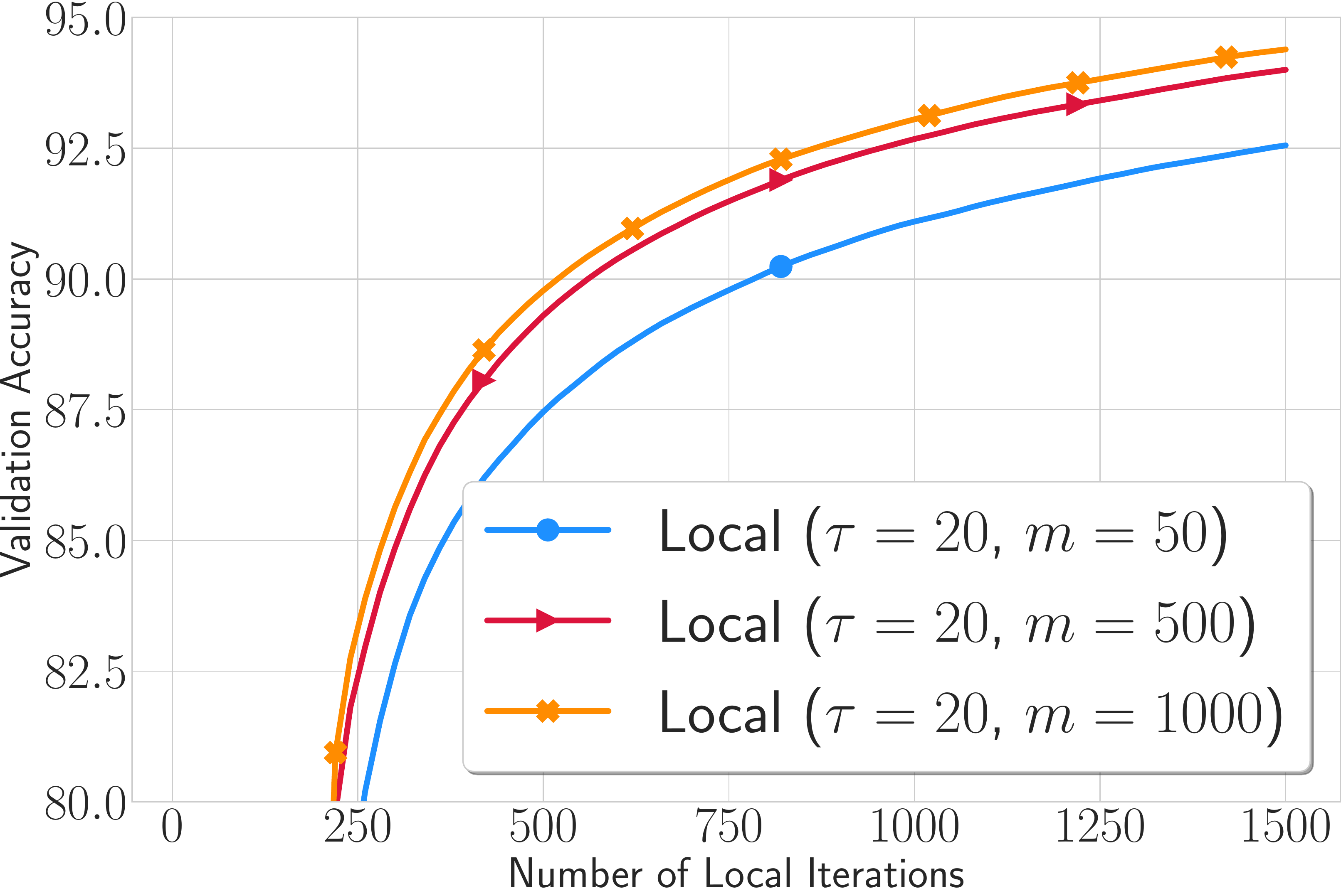}
    % \caption{Caption}
    \label{fig:my_label2}
    \end{subfigure}
    
    \begin{subfigure}[b]{0.234\textwidth}
    \centering
    \includegraphics[width=\textwidth]{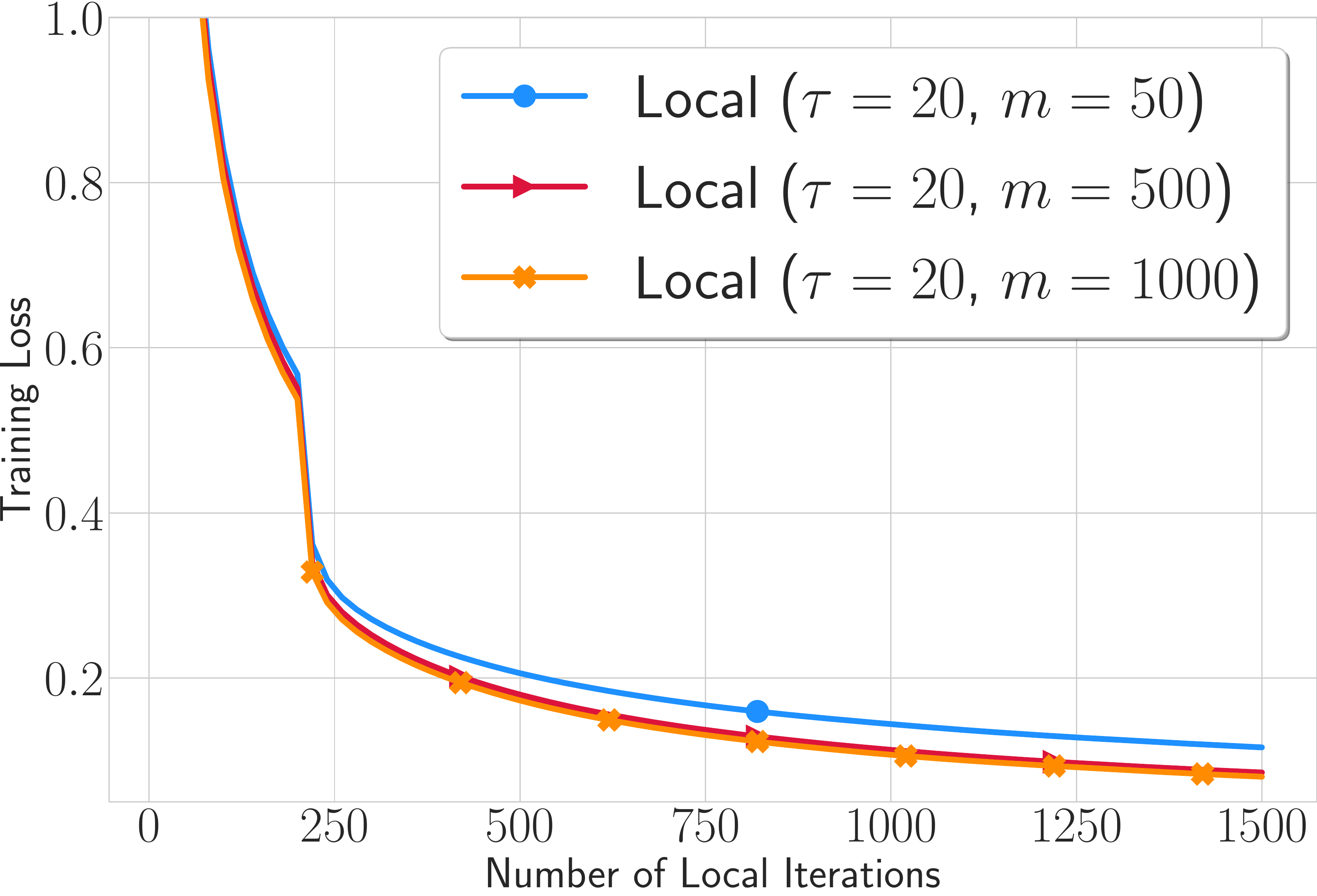}
    \caption{Homogeneous Data Distribution}
    \label{fig:my_label3}
    \end{subfigure}
    % \hspace{1cm}
    \begin{subfigure}[b]{0.234\textwidth}
    \centering
    \includegraphics[width=\textwidth]{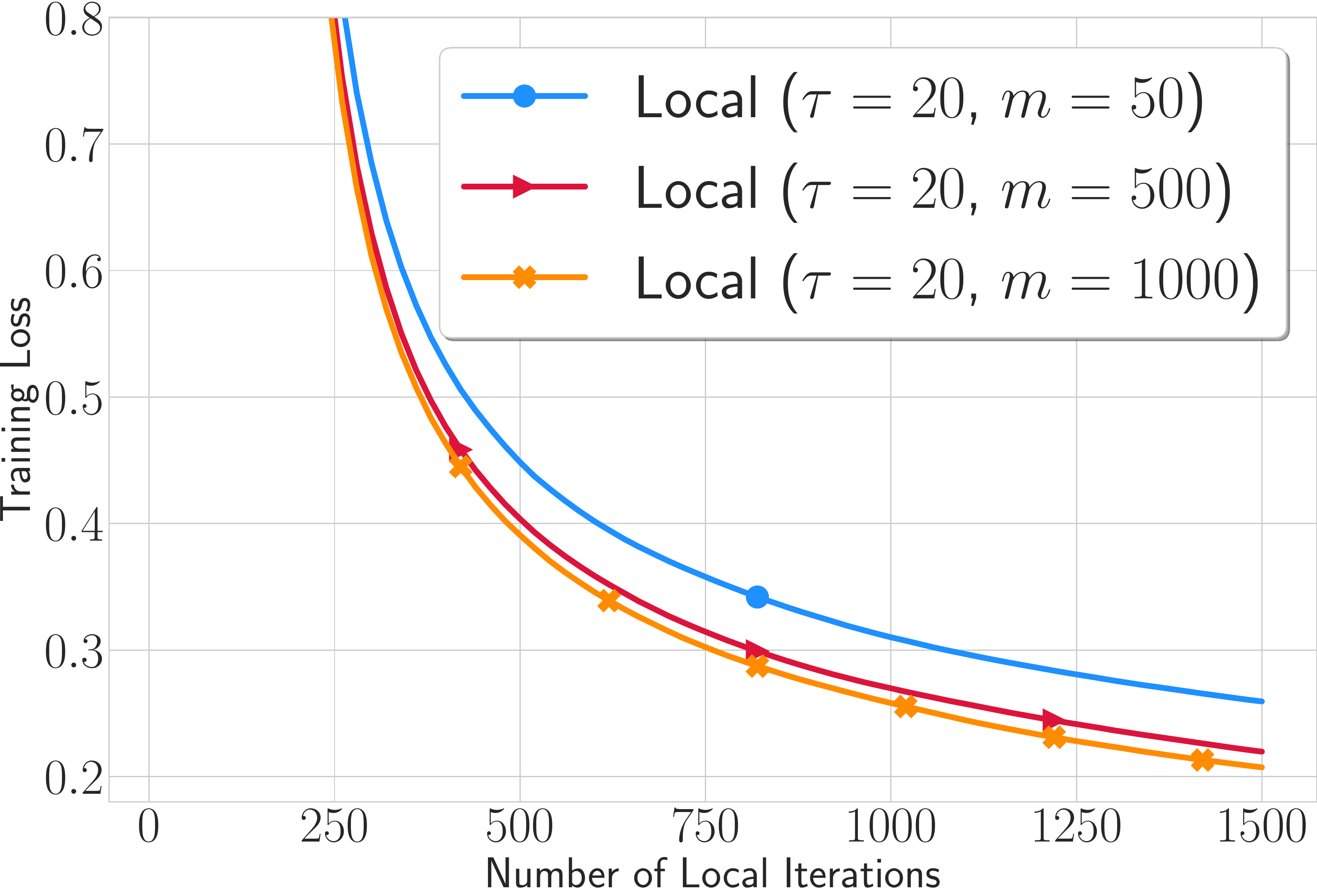}
    \caption{Heterogeneous Data Distribution}
    \label{fig:my_label4}
    \end{subfigure}
    \caption{Comparing the effect of model size using Local SGD on homogeneous and heterogeneous data distribution. By changing the model size from $m=50$ to $m=1000$, the model converges faster. In heterogeneous setting the increase in the model size has more impact on the convergence rate than the homogeneous setting. }
    \label{fig:change_m} \vspace{-0.5cm}
\end{figure}

\noindent\textbf{Effects of different model sizes $m$.} We firstly train the model using Local SGD with the same synchronization gap and different number of hidden neurons, $m$. Figure~\ref{fig:change_m} shows the results of this experiment on models with different hidden layer's size in homogeneous and heterogeneous settings. As it can be seen in both cases, the model with higher model size can achieve better final accuracy. This phenomenon has more impact in the  heterogeneous data distribution compared to homogeneous setting.

 \begin{figure*}
    \centering
    \begin{subfigure}[b]{0.35\textwidth}
    \centering
    \includegraphics[width=\textwidth]{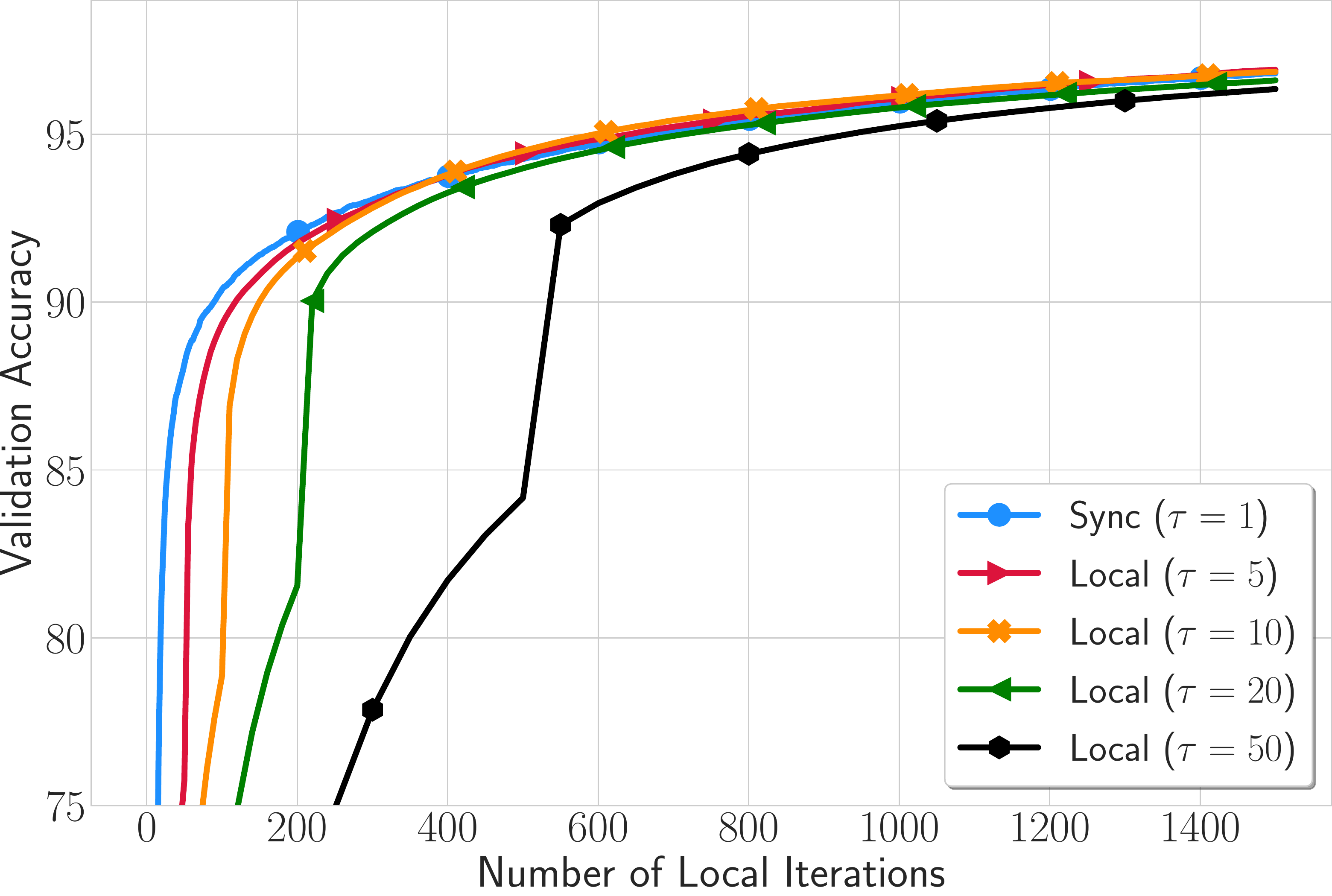}
    % \caption{Caption}
    \label{fig:my_label1}
    \end{subfigure}
    \hspace{1cm}
    \begin{subfigure}[b]{0.35\textwidth}
    \centering
    \includegraphics[width=\textwidth]{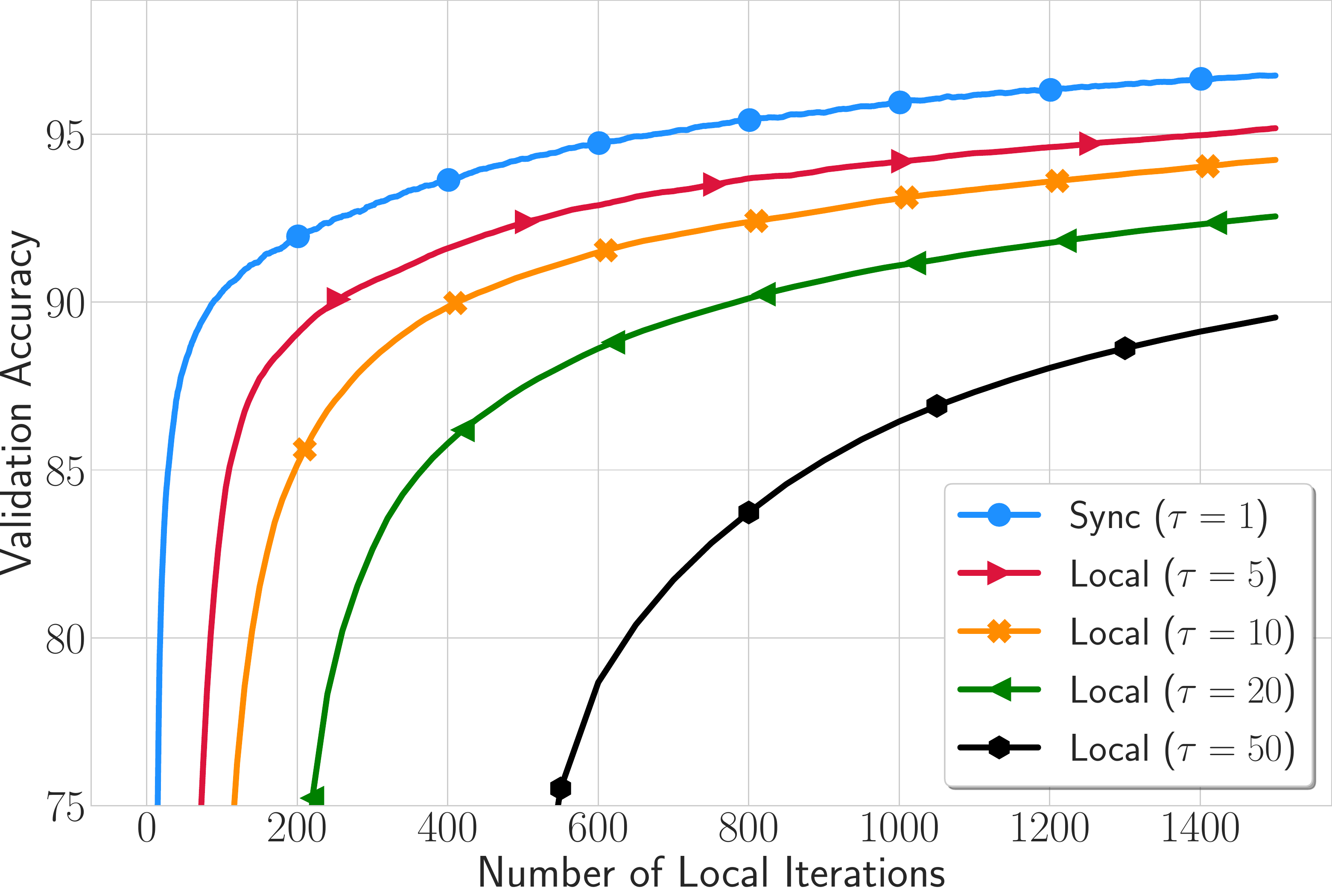}
    % \caption{Caption}
    \label{fig:my_label2}
    \end{subfigure}
    
    \begin{subfigure}[b]{0.35\textwidth}
    \centering
    \includegraphics[width=\textwidth]{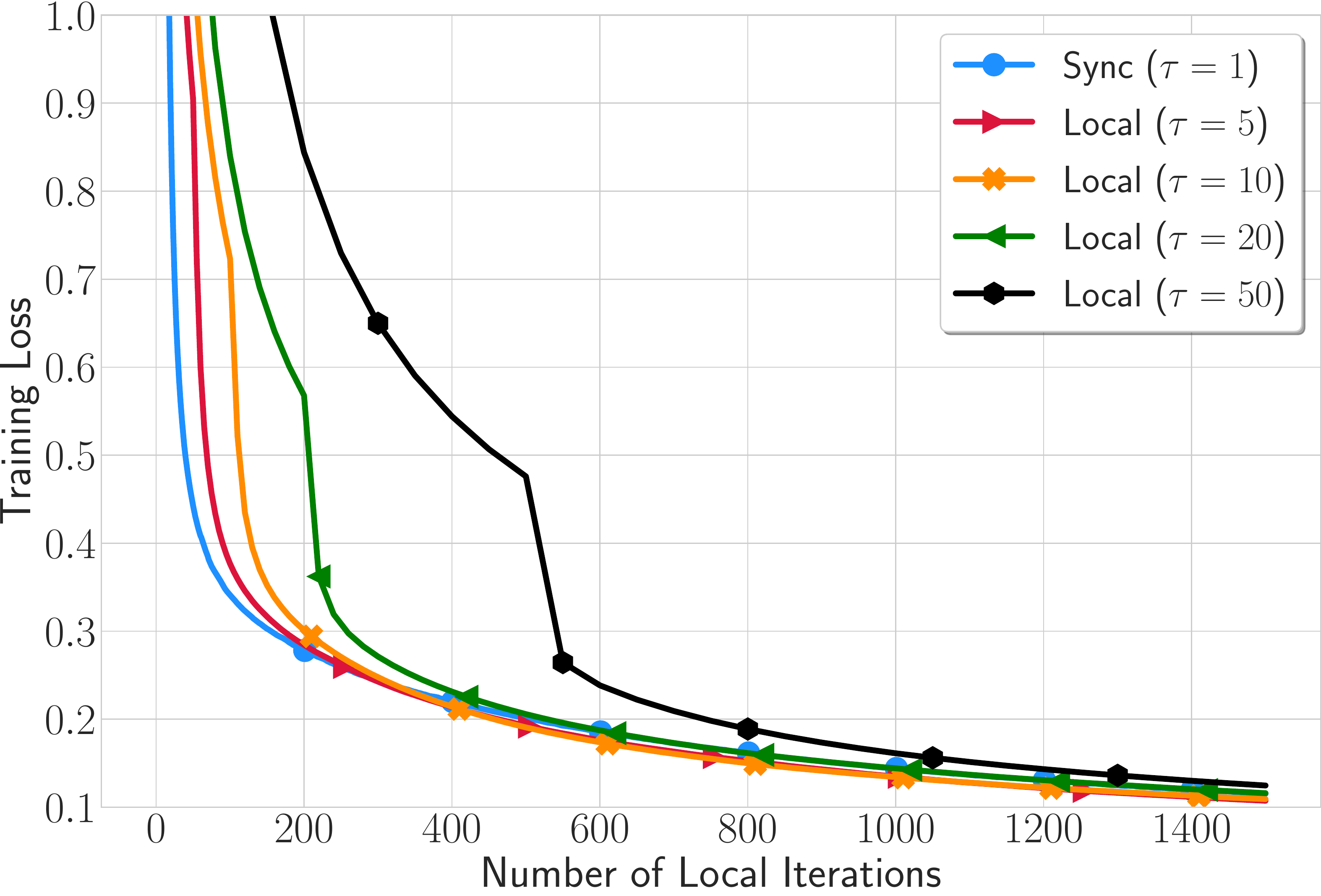}
    \caption{Homogeneous Data Distribution}
    \label{fig:my_label3}
    \end{subfigure}
    \hspace{1cm}
    \begin{subfigure}[b]{0.35\textwidth}
    \centering
    \includegraphics[width=\textwidth]{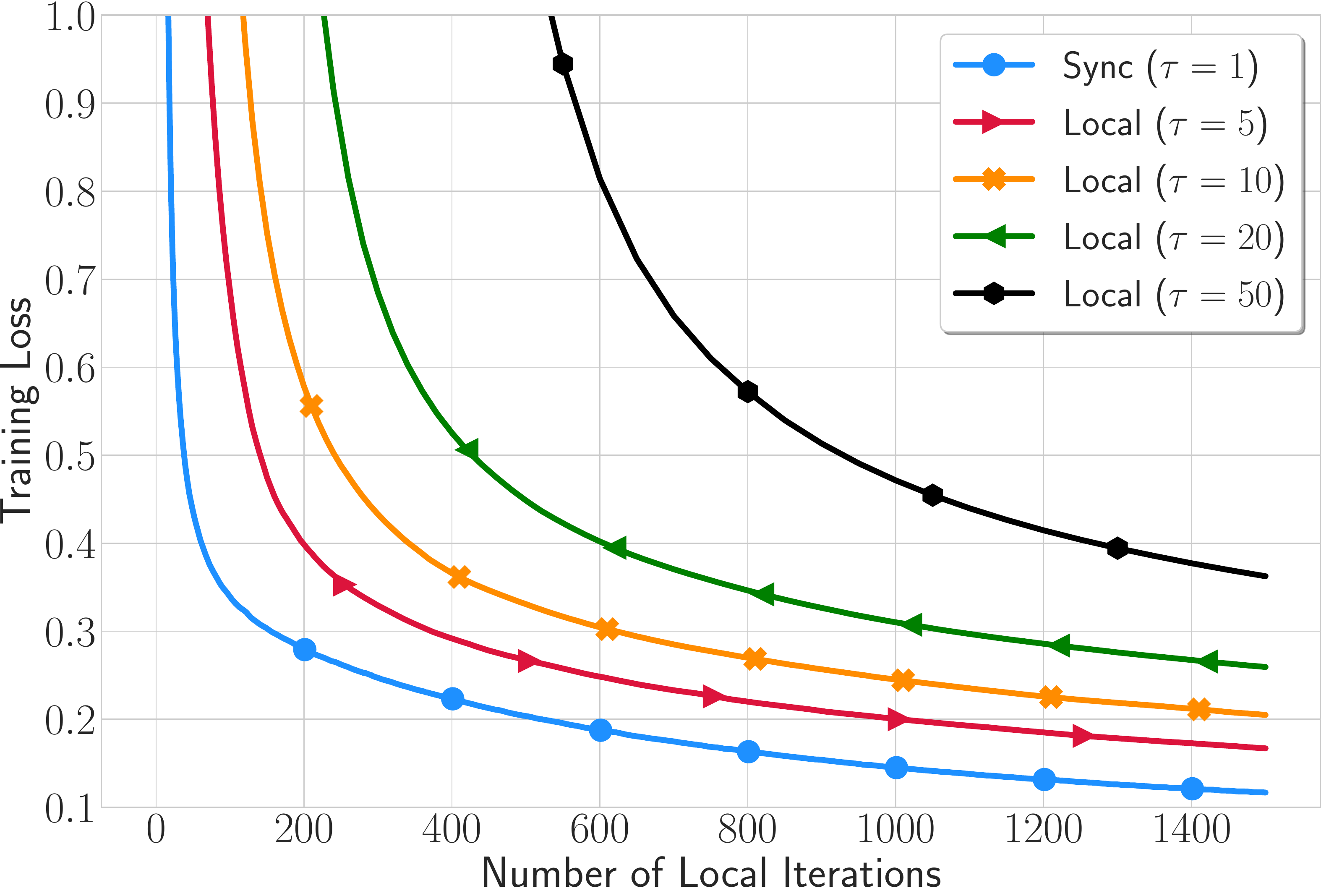}
    \caption{Heterogeneous Data Distribution}
    \label{fig:my_label4}
    \end{subfigure}
    \caption{Comparing the effect of synchronization gap (i.e., number of local updates $\tau$) on the model convergence. In this experiment the model size is fixed on $m=50$. }
    \label{fig:change_tau} \vspace{-0.5cm}
\end{figure*}

\noindent\textbf{Effects of different synchronization gap $\tau$.} Now, we fix the model size $m=50$ and change the synchronization gap $\tau$. We do the comparison between fully synchronous SGD and Local SGD with $\tau=5,10,20,50$. The results in Figure~\ref{fig:change_tau} shows that in both homogeneous and heterogeneous settings, the convergence rate becomes slower when synchronization gap increases. However, in the heterogeneous setting, increasing $\tau$ will decrease the convergence speed more significantly.

\begin{figure}
    \centering\vspace{-0.5cm}
    \begin{subfigure}[b]{0.35\textwidth}
    \centering
    \includegraphics[width=\textwidth]{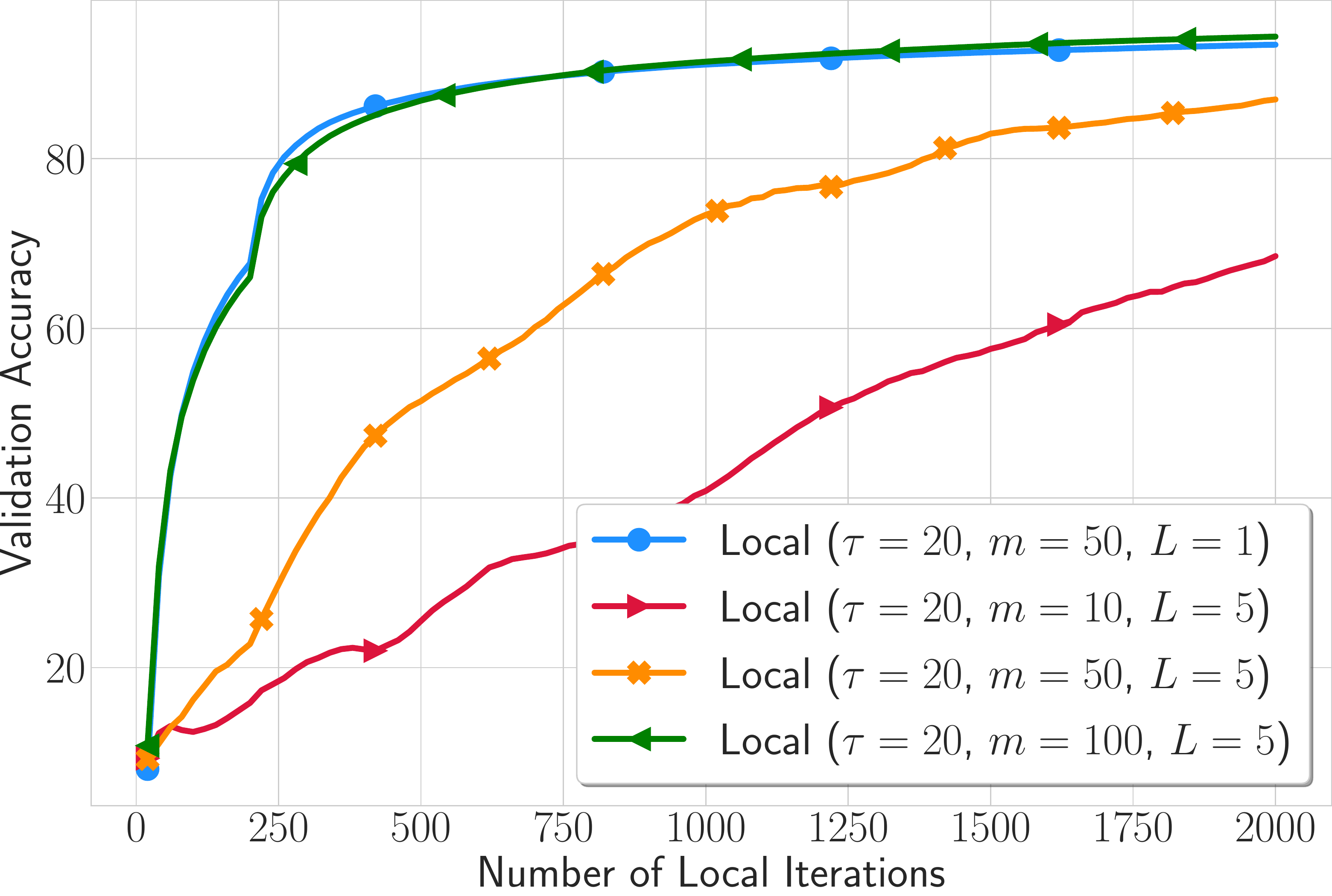}
    % \caption{Caption}
    \label{fig:acc_l5}
    \end{subfigure}
    \hfill
    \begin{subfigure}[b]{0.35\textwidth}
    \centering
    \includegraphics[width=\textwidth]{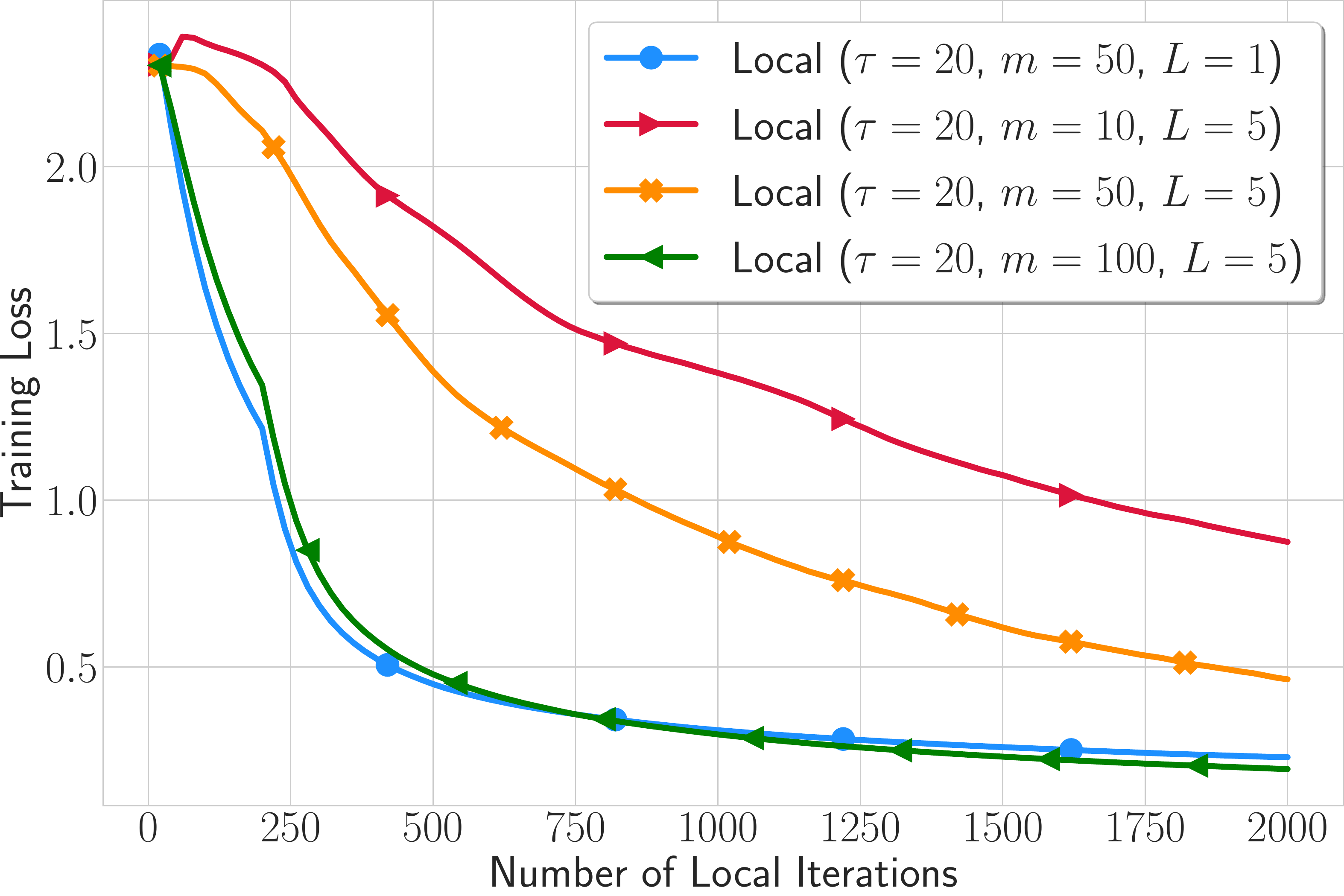}
    % \caption{Caption}
    \label{fig:loss_l5}
    \end{subfigure}
    \caption{The effect of number of layer $L$. If we increase the number of layers to 5, compared to single layer, we need more neurons ($m$) to converge with the same rate as single layer.}\label{fig:l5} \vspace{-0.5cm}
\end{figure}

\noindent\textbf{Effects of number of layers $L$.} When we increase the number of layers $L$, based on condition of $m$ in Theorem~\ref{theorem: local SGD}, we need to increase the number of neurons as well to achieve the same rate. For instance, Figure~\ref{fig:l5} shows the convergence rate of models with $L=5$ and various $m\in\{10,50,100\}$, compared with the single layer model with $m=50$. If we use the same number of neurons as the single layer (i.e. $m=10$ and $L=5$), it is evident that the model performs poorly. By increasing the number of neurons per layer, we can see that $m=100$ can make 5-layer model achieve the same performance of the single layer model, which has $10\times$ more neurons and more than $3\times$ bigger in terms of parameter size. This is consistent with Theorem~\ref{theorem: local SGD}, as increasing the number of layers requires significantly more neurons \textit{per layer} compared to single layer counterpart to guarantees linear convergence rate.

\section{Discussion and Future Works}\label{sec: discussion}
In this paper, we proved
that both Local GD and Local SGD that are originally proposed for communication efficient  training of deep neural networks can achieve global minima of the training loss for over-parameterized
deep ReLU networks. We make the first theoretical trial on the analysis of Local (S)GD on training Deep ReLU networks with multiple layers, but we do not claim that our results, e.g., number of required neurons and dependency on the number of layers, are optimal in any sense. 

A number of future works/improvements are still exciting to explore:

\noindent\textbf{Tightening the condition on the number of neurons.}~In the bounds obtained for both Local GD and SGD, the required   number of neurons to has a heavy dependency on the number training samples $n$, which is  worse than the single machine case. We are aware of some recent works~\cite{noy2021convergence,nguyen2020global,nguyen2021proof} that demonstrate significantly reduced   number of required neurons, and  we believe incorporating their results can also improve our theory to entail tighter bounds.

\noindent\textbf{Extension of analysis to other federated optimization methods.} To further reduce the harm caused by multiple local updates, a line of recent studies proposed alternative methods to reduce the  local model deviation~\cite{karimireddy2019scaffold,yuan2020federated}. Establishing the  convergence of these variants on deep non-smooth networks is another valuable research direction.

\noindent\textbf{Extension to other neural network architectures.} In this paper we only consider simple ReLU forward feed neural network, but as shown in the prior works~\cite{allen2019convergence,zou2020global}, single machine SGD can optimize more complicated neural network like CNN, ResNet or RNN as well. Hence, one natural future work is to extend our analysis on Local SGD to those neural network models.
\section*{Acknowledgement}
This work was supported in part by NSF grant 1956276.
 
\clearpage 
 
\bibliographystyle{plain}
\bibliography{references}

\clearpage 
\onecolumn
 \clearpage
 \appendix

\section{Proof of Theorem~\ref{theorem: local GD}  (Local GD)}\label{app: GD}
In this section  we  present the proof of convergence rate of Local GD (Theorem~\ref{theorem: local GD}).  Similar to analysis of Local SGD for smooth objectives~\cite{stich2018local,khaled2020tighter}, we start from pertubed virtual iterates analysis. We let $\mathbf{W} (t) = \frac{1}{K}\sum_{i=1}^K \mathbf{W}^{(i)}(t)$ denote the virtual averaged iterates. Before providing the proof, let us first introduce some useful lemmas. 
\subsection{Proof of Technical Lemma}
The following lemma is from Allen-Zhu et al's seminal work~\cite{allen2019convergence}, which characterizes the forward perturbation property of deep ReLU network:
\begin{lemma}[Allen et al~\cite{allen2019convergence}]\label{lemma:perturbation}
Consider a weight matrices $\tilde{\mathbf{W}} $, $\mathbf{W}$ such that $\tilde{\mathbf{W}}, \mathbf{W}\in\mathcal{B}(\mathbf{W}(0),\omega)$, with probability at least $1-\exp(-O(m\omega^{2/3}))$, the following facts hold:
\begin{align*} 
    &\|f_{j,l-1} - f_{j,l-1}(0)\| \leq O\left(\omega L^{5/2} \sqrt{\log m}\right),\\
    &\|f_{j,l-1}\| \leq O(1),\\
    &\left\| \boldsymbol{v}^\top  \mathbf{V}( {\mathbf{D}}_{j,L} {\mathbf{W}}_L\cdots {\mathbf{D}}_{j,l+1} {\mathbf{W}}_{l+1}  {\mathbf{D}}_{j,l}-\tilde{\mathbf{D}}_{j,L}\tilde{\mathbf{W}}_L\cdots\tilde{\mathbf{D}}_{j,l+1}\tilde{\mathbf{W}}_{l+1} \tilde{\mathbf{D}}_{j,l}) \right\|  \leq O\left(\omega^{1/3} L^2 \sqrt{m\log m/d}\right)\cdot \| \boldsymbol{v} \|,\\
    &\|f_{j,l-1} - \tilde{f}_{j,l-1}\| \leq O\left(L^{3/2} \|\mathbf{W}-\tilde{\mathbf{W}}\|_2\right),\\
    &\left\| \boldsymbol{v}^\top \mathbf{V}( {\mathbf{D}}_{j,L} {\mathbf{W}}_L\cdots {\mathbf{D}}_{j,l+1} {\mathbf{W}}_{l+1}  {\mathbf{D}}_{j,l}  \right\|  \leq O\left(\sqrt{m/d}\right)\cdot \| \boldsymbol{v} \|,
\end{align*} 
where $\boldsymbol{v}$ is arbitrary vector, $f_{j,l-1}(0) =  \sigma (\mathbf{W}_l(0) \sigma(\mathbf{W}_{l-1}(0)\cdots \sigma(\mathbf{W}_1(0) \bm{x}_j)))$.
\end{lemma}

The following lemma establishes a bound on the deviation between local models and (virtual) averaged global model in terms of global loss.
 
\begin{lemma}\label{lemma: weight gap GD}
For Local GD, let $t_c$ denote  the latest communication stage of before iteration $t$. If the condition that $L_i(\mathbf{W}^{(i)}(t) ) \leq L_i(\mathbf{W}(t_c) )$ for any $t_c \leq t\leq t_c + \tau -1$ holds, then the following statement holds true for $t_c \leq t \leq t_c + \tau-1$: 
\begin{align*}
 \left\| \mathbf{W}^{(i)}(t) -  \mathbf{W}(t_c) \right\|^2_{\mathrm{F}}  
    & \leq     O\left((\eta^2 \tau^2   + \eta^2 \tau)\frac{mn}{d}\right) L_i(\mathbf{W}(t_c) ). 
\end{align*}
where $K$ is the number of devices, $\tau$ is the number of local updates between two consecutive rounds of synchronization, $n$ is the size of each local data shard, and $m$ is the number of neurons in hidden layer.
\begin{proof}
According to updating rule we have:
\begin{align*}
    \left\| \mathbf{W}^{(i)}(t+1) -  \mathbf{W}(t_c) \right\|^2_{\mathrm{F}} =   &\left\| \mathbf{W}^{(i)}(t) - \eta \nabla L_i(\mathbf{W}^{(i)}(t))- \mathbf{W}(t_c) \right\|^2_{\mathrm{F}}  \\
    & = \left\| \mathbf{W}^{(i)}(t) - \mathbf{W}(t_c) \right\|^2_{\mathrm{F}} - 2\eta \left \langle \nabla L_i(\mathbf{W}^{(i)}(t)), \mathbf{W}^{(i)}(t) - \mathbf{W}(t_c) \right\rangle\\
    &\quad  + \eta^2 \|\nabla L_i(\mathbf{W}^{(i)}(t))\|^2_\mathrm{F}\\
    & = \left\| \mathbf{W}^{(i)}(t) - \mathbf{W}(t_c) \right\|^2_{\mathrm{F}} + 2\eta \left \langle \nabla L_i(\mathbf{W}^{(i)}(t)),  \eta \sum_{t' = t_c}^{t-1} \nabla L_i(\mathbf{W}^{(i)}(t)) \right\rangle\\
    &\quad  + \eta^2 \|\nabla L_i(\mathbf{W}^{(i)}(t))\|^2_\mathrm{F}\\
     & = \left\| \mathbf{W}^{(i)}(t) - \mathbf{W}(t_c) \right\|^2_{\mathrm{F}} + 2\eta^2 \tau \left \langle \nabla L_i(\mathbf{W}^{(i)}(t)),   \frac{1}{\tau} \sum_{t' = t_c}^{t-1} \nabla L_i(\mathbf{W}^{(i)}(t)) \right\rangle\\
    &\quad  + \eta^2 \|\nabla L_i(\mathbf{W}^{(i)}(t))\|^2_\mathrm{F}. 
\end{align*}
Applying the identity $\langle \bm{a}, \bm{b} \rangle = \frac{1}{2}\|\bm{a}\|^2+\frac{1}{2}\|\bm{b}\|^2 - \frac{1}{2}\|\bm{a}-\bm{b}\|^2$ on the cross term yields:
 \begin{align*}
    &\left\| \mathbf{W}^{(i)}(t+1) -  \mathbf{W}(t_c) \right\|^2_{\mathrm{F}}\\
    & =   \left\| \mathbf{W}^{(i)}(t) - \eta \nabla L_i(\mathbf{W}^{(i)}(t))- \mathbf{W}(t_c) \right\|^2_{\mathrm{F}}  \\
     & \leq \left\| \mathbf{W}^{(i)}(t) - \mathbf{W}(t_c) \right\|^2_{\mathrm{F}} +  \eta^2 (t-t_c) \left(\left \| \nabla L_i(\mathbf{W}^{(i)}(t))\right\|^2+\left \|   \frac{1}{(t-t_c)} \sum_{t' = t_c}^{t-1} \nabla L_i(\mathbf{W}^{(i)}(t)) \right\|^2 \right)\\
    &\quad  + \eta^2 \|\nabla L_i(\mathbf{W}^{(i)}(t))\|^2_\mathrm{F} \\
    & \leq \left\| \mathbf{W}^{(i)}(t) - \mathbf{W}(t_c) \right\|^2_{\mathrm{F}} +  \eta^2 (t-t_c) \left(\left \| \nabla L_i(\mathbf{W}^{(i)}(t))\right\|^2+ \frac{1}{t-t_c} \sum_{t' = t_c}^{t-1}\left \|   \nabla L_i(\mathbf{W}^{(i)}(t)) \right\|^2 \right)\\
    &\quad  + \eta^2 \|\nabla L_i(\mathbf{W}^{(i)}(t))\|^2_\mathrm{F}.
\end{align*}
Plugging the gradient upper bound from Lemma~\ref{lemma:grad bounds} yields:
 \begin{align*}
    &\left\| \mathbf{W}^{(i)}(t+1) -  \mathbf{W}(t_c) \right\|^2_{\mathrm{F}}\\
    &=   \left\| \mathbf{W}^{(i)}(t) - \eta \nabla L_i(\mathbf{W}^{(i)}(t))- \mathbf{W}(t_c) \right\|^2_{\mathrm{F}}  \\ 
    & \leq \left\| \mathbf{W}^{(i)}(t) - \mathbf{W}(t_c) \right\|^2_{\mathrm{F}} +  \eta^2 (t-t_c) \left(O\big(\frac{mn}{d}\big) L_i(\mathbf{W}^{(i)}(t) ) + \frac{1}{t-t_c} \sum_{t' = t_c}^{t-1}O\big(\frac{mn}{d}\big) L_i(\mathbf{W}^{(i)}(t') )  \right)\\
    &\quad  + \eta^2 O\big(\frac{mn}{d}\big) L_i(\mathbf{W}^{(i)}(t) )  .
\end{align*}
Since we assume $L_i(\mathbf{W}^{(i)}(t) ) \leq L_i(\mathbf{W}(t_c) )$ for any $t_c \leq t\leq t_c + \tau -1$, so we have:
 \begin{align*}
    \left\| \mathbf{W}^{(i)}(t+1) -  \mathbf{W}(t_c) \right\|^2_{\mathrm{F}} &=  \left\| \mathbf{W}^{(i)}(t) - \eta \nabla L_i(\mathbf{W}^{(i)}(t))- \mathbf{W}(t_c) \right\|^2_{\mathrm{F}}  \\ 
    & \leq \left\| \mathbf{W}^{(i)}(t) - \mathbf{W}(t_c) \right\|^2_{\mathrm{F}} +  \eta^2 \tau \left(O\big(\frac{mn}{d}\big) L_i(\mathbf{W}^{(i)}(t_c) ) +  O\big(\frac{mn}{d}\big) L_i(\mathbf{W}^{(i)}(t_c) )  \right)\\
    &\quad  + \eta^2 O\big(\frac{mn}{d}\big) L_i(\mathbf{W}^{(i)}(t_c) ) . 
\end{align*}
Doing the telescoping sum from $t+1$ to $t_c$ will conclude the proof:
 \begin{align*}
    \left\| \mathbf{W}^{(i)}(t+1) -  \mathbf{W}(t_c) \right\|^2_{\mathrm{F}}  
    & \leq     O\left((\eta^2 \tau^2   + \eta^2 \tau)\frac{mn}{d}\right) L_i(\mathbf{W}^{(i)}(t_c) ). 
\end{align*} 
\end{proof} 
\end{lemma}

The next lemma is the key result in our proof, which characterizes the semi gradient Lipschitzness property of ReLU neural network.     
 \begin{lemma}[\sffamily{Semi-gradient Lipschitzness}]\label{lm: gradient gap} For Local GD, at any iteration $t$, if $\mathbf{W} , \tilde{\mathbf{W}}  \in \mathcal{B}(\bm{W}(0),\omega)$, then with probability at least $1-\exp\left(-\Omega(m\omega^{2/3})\right)$, the following statement holds true:
\begin{align*}
    \frac{1}{K}\sum_{i=1}^K \left\| \nabla_{\mathbf{W}}L_i(\mathbf{W})- \nabla_{\tilde{\mathbf{W}} }L_i  (\tilde{\mathbf{W}})\right\|^2_{\mathrm{F}} &\leq O\left( \frac{m L^4}{d} \left\|\mathbf{W} - \tilde{\mathbf{W}}\right\|_2^2\right)+O\left(\frac{\omega^{2/3} L^5 {m\log m }}{d}+ \frac{\omega^2 L^{6 } {m \log m}}{d}\right) L(\tilde{\mathbf{W}}),
\end{align*}
where $K$ is the number of devices, $\tau$ is the number of local updates between two consecutive rounds of synchronization, $n$ is the size of each local data shard, and $m$ is the number of neurons in hidden layer.
\begin{proof}
Observe that:
\begin{align*}
   &\frac{1}{K}\sum_{i=1}^K \left\|\nabla_{\mathbf{W}}L_i(\mathbf{W})- \nabla_{\tilde{\mathbf{W}} }L_i  (\tilde{\mathbf{W}})\right\|^2_{\mathrm{F}}   = \frac{1}{K}\sum_{i=1}^K \sum_{l=1}^L \left\|\nabla_{\mathbf{W}_l}L_i(\mathbf{W})- \nabla_{\tilde{\mathbf{W}}_l }L_i  (\tilde{\mathbf{W}})\right\|^2_{\mathrm{F}}\nonumber.
\end{align*} 
Let $f_j^{(i)} = \mathbf{V}\mathbf{D}^{(i)}_{j,L}\mathbf{W}^{(i)}_L\cdots\mathbf{D}_{j,1}\mathbf{W}^{(i)}_{1} \bm{x}_j$ and $\mathbf{L}^{(i)}_j = f_j^{(i)} - \bm{y}_j$. Now we examine the difference of the gradients:
\begin{align*}
 &\nabla_{\mathbf{W}_l}L_i(\mathbf{W})- \nabla_{\tilde{\mathbf{W}}_l }L_i  (\tilde{\mathbf{W}}) \\
 &= \frac{1}{n} \sum_{j=1}^n \left[ (\mathbf{L}^{(i)}_j{}^\top \mathbf{V}\mathbf{D}_{j,L}\mathbf{W}_L\cdots\mathbf{D}_{j,l+1}\mathbf{W}_{l+1} \mathbf{D}_{j,l})^\top (f_{j,l-1}^{(i)}){}^\top - ({\tilde{\mathbf{L}}^{(i)}_j}{}^\top \mathbf{V}\tilde{\mathbf{D}}_{j,L}\tilde{\mathbf{W}}_L\cdots\tilde{\mathbf{D}}_{j,l+1}\tilde{\mathbf{W}}_{l+1} \tilde{\mathbf{D}}_{j,l})^\top (\tilde{f}_{j,l-1}^{(i)})^\top \right]\\
 & = \frac{1}{n} \sum_{j=1}^n \left[ (({\mathbf{L}^{(i)}_j}^\top-{\tilde{\mathbf{L}}^{(i)}_j}{}^\top) \mathbf{V}\mathbf{D}_{j,L}\mathbf{W}_L\cdots\mathbf{D}_{j,l+1}\mathbf{W}_{l+1} \mathbf{D}_{j,l})^\top  (f_{j,l-1}^{(i)}){}^\top    \right]\\
 & \quad +  \frac{1}{n} \sum_{j=1}^n \left[ \left({\tilde{\mathbf{L}}^{(i)}_j}{}^\top \mathbf{V}( {\mathbf{D}}_{j,L} {\mathbf{W}}_L\cdots {\mathbf{D}}_{j,l+1} {\mathbf{W}}_{l+1}  {\mathbf{D}}_{j,l}-\tilde{\mathbf{D}}_{j,L}\tilde{\mathbf{W}}_L\cdots\tilde{\mathbf{D}}_{j,l+1}\tilde{\mathbf{W}}_{l+1} \tilde{\mathbf{D}}_{j,l})\right)^\top ( {f}_{j,l-1}^{(i)}){}^\top  \right]\\
 & \quad +  \frac{1}{n} \sum_{j=1}^n \left[ ({\tilde{\mathbf{L}}^{(i)}_j}{}^\top \mathbf{V}\tilde{\mathbf{D}}_{j,L}\tilde{\mathbf{W}}_L\cdots\tilde{\mathbf{D}}_{j,l+1}\tilde{\mathbf{W}}_{l+1} \tilde{\mathbf{D}}_{j,l})^\top ( {f}_{j,l-1}^{(i)}-\tilde{f}_{j,l-1}^{(i)}){}^\top \right].
\end{align*} 
According to Lemma~\ref{lemma:perturbation} we know the following facts:
\begin{align*}
    &\|f_{j,l-1}^{(i)} - \tilde{f}_{j,l-1}^{(i)}\| \leq O\left(\omega L^{5/2} \sqrt{\log m}\right),\\
    &\|f_{j,l-1}^{(i)}\| \leq O(1),\\
    &\left\| \left(\tilde{\mathbf{L}}^{(i)}_j{}^\top \mathbf{V}( {\mathbf{D}}_{j,L} {\mathbf{W}}_L\cdots {\mathbf{D}}_{j,l+1} {\mathbf{W}}_{l+1}  {\mathbf{D}}_{j,l}-\tilde{\mathbf{D}}_{j,L}\tilde{\mathbf{W}}_L\cdots\tilde{\mathbf{D}}_{j,l+1}\tilde{\mathbf{W}}_{l+1} \tilde{\mathbf{D}}_{j,l})\right)^\top \right\|  \leq O\left(\omega^{1/3} L^2 \sqrt{m\log m/d}\right)\cdot \|\tilde{\mathbf{L}}^{(i)}_j\|, \\
    & \left\| \mathbf{L}^{(i)}_j{}^\top-\tilde{\mathbf{L}}^{(i)}_j{}^\top \right\|_\mathrm{F} = \left\| \mathbf{V} \sigma(\mathbf{W}_L\cdots \sigma (\mathbf{W}_1 \bm{x}_j))-\mathbf{V} \sigma(\tilde{\mathbf{W}}_L\cdots \sigma (\tilde{\mathbf{W}}_1 \bm{x}_j)) \right\|  \leq O\left(L^{3/2}\left\|\mathbf{W} - \tilde{\mathbf{W}}\right\|_2\right),\\
    & \left\| (\mathbf{L}^{(i)}_j{}^\top-\tilde{\mathbf{L}}^{(i)}_j{}^\top) \mathbf{V}\mathbf{D}_{j,L}\mathbf{W}_L\cdots\mathbf{D}_{j,l+1}\mathbf{W}_{l+1} \mathbf{D}_{j,l})^\top\right\| \leq   O\left(\sqrt{m/d}\cdot L^{3/2} \left\|\mathbf{W} - \tilde{\mathbf{W}}\right\|_2\right). 
\end{align*}

So we have the following bound for Frobenius norm: 
\begin{align*}
    &\left\|\nabla_{\mathbf{W}_l}L_i(\mathbf{W})- \nabla_{\tilde{\mathbf{W}}_l }L_i  (\tilde{\mathbf{W}})\right\|_{\mathrm{F}}^2\\
    &\leq \left(O\left(\sqrt{m/d}\cdot L^{3/2} \left\|\mathbf{W} - \tilde{\mathbf{W}}\right\|_2\right)+O\left(\omega^{1/3} L^2 \sqrt{m\log m/d}\right)\cdot \|\tilde{\mathbf{L}}^{(i)}_j\|+O\left(\omega L^{5/2}\sqrt{m \log m/d}   \right)\|\tilde{\mathbf{L}}^{(i)}_j\|\right)^2\\
    &\leq O\left( {m/d}\cdot L^{3} \left\|\mathbf{W} - \tilde{\mathbf{W}}\right\|_2^2\right)+O\left(\omega^{2/3} L^4 {m\log m/d}+\omega^2 L^{5 } {m \log m/d}   \right)\|\tilde{\mathbf{L}}^{(i)}_j\|^2.
\end{align*} 

Hence we can conclude the proof:
\begin{align*}
   &\frac{1}{K}\sum_{i=1}^K \left\|\nabla_{\mathbf{W}}L_i(\mathbf{W})- \nabla_{\tilde{\mathbf{W}} }L_i  (\tilde{\mathbf{W}})\right\|^2_{\mathrm{F}}   = \frac{1}{K}\sum_{i=1}^K \sum_{l=1}^L \left\|\nabla_{\mathbf{W}_l}L_i(\mathbf{W})- \nabla_{\tilde{\mathbf{W}}_l }L_i  (\tilde{\mathbf{W}})\right\|^2_{\mathrm{F}}\nonumber\\ 
   & \leq O\left( \frac{m L^4}{d} \left\|\mathbf{W} - \tilde{\mathbf{W}}\right\|_2^2\right)+O\left(\frac{\omega^{2/3} L^5 {m\log m }}{d}+ \frac{\omega^2 L^{6 } {m \log m}}{d}\right) L(\tilde{\mathbf{W}}).
\end{align*} 
\end{proof}
\end{lemma}

 \begin{lemma}\label{lm: shrink of local loss GD}
 For Local GD, at any iteration $t$ in between two communication rounds: $t_c \leq t\leq t_c + \tau -1$ and $i \in [K]$, if $\mathbf{W}^{(i)}(t) \in \mathcal{B}(\bm{W}(0),\omega)$, then with probability at least $1-\exp\left(-\Omega(m\omega^{2/3})\right)$, the following statement holds true:
 \begin{align*}
     L_i(\mathbf{W}^{(i)}(t)) \leq  L_i(\mathbf{W}^{(i)}(t-1)) \leq \cdots \leq L_i(\mathbf{W}^{(i)}(t_c)).
 \end{align*} 
 \begin{proof}
 According to updating rule and the semi smoothness property:
 \begin{align*} 
&L_i(\mathbf{W}^{(i)}(t))\le L_i( \mathbf{W}^{(i)}(t-1)) + \left[\left\langle \nabla L_i( \mathbf{W}^{(i)}(t-1)),   \mathbf{W}^{(i)}(t)-  \mathbf{W}^{(i)}(t-1)\right \rangle \right]\notag\\
&\quad+ C'\sqrt{L_i(\mathbf{W}^{(i)}(t))}\cdot\frac{\omega^{1/3}L^2\sqrt{m\log(m)}}{\sqrt{d}}\cdot \|\mathbf{W}^{(i)}(t)-  \mathbf{W}^{(i)}(t-1)\|_2 + \frac{C''L^2m}{d}\|\mathbf{W}^{(i)}(t)-  \mathbf{W}^{(i)}(t-1)\|^2_2\\
&\le L( \mathbf{W}^{(i)}(t-1)) -\eta  \left\langle \nabla L( \mathbf{W}^{(i)}(t-1)),     \nabla  L_i(\mathbf{W}^{(i)}(t-1))  \right \rangle \notag\\
&\quad+\eta  C'\sqrt{ L_i(\mathbf{W}^{(i)}(t-1))}\cdot\frac{\omega^{1/3}L^2\sqrt{m\log(m)}}{\sqrt{d}}\cdot \left\|   \nabla  L_i(\mathbf{W}^{(i)}(t-1))\right \|_2   + \eta^2\frac{C''L^2m}{d}\mathbb{E}\left\|   \nabla  L_i(\mathbf{W}^{(i)}(t-1))\right\|^2_2  \nonumber\\
& \stackrel{\text{\ding{192}}}{\leq} \left(1-\Omega\left({\frac{\eta \tau m \phi}{dn^2}}\right)\right)L_i(\mathbf{W}^{(i)}(t-1)),
\end{align*}
where in \ding{192} we plug in the gradient upper bound from Lemma~\ref{lemma:grad bounds}.
According to our choice of $\eta$, we can conclude that \begin{align*}
     L_i(\mathbf{W}^{(i)}(t)) \leq  L_i(\mathbf{W}^{(i)}(t-1)) \leq \cdots \leq L_i(\mathbf{W}^{(i)}(t_c)).
 \end{align*} 
 \end{proof}
 
 \end{lemma}

 \subsection{Proof of Theorem~\ref{theorem: local GD}}
With the key lemmas in place, we now  prove Theorem~\ref{theorem: local GD} by induction. Assume the following induction hypotheses hold for $h \leq t$:
 \begin{align}
      &\textbf{(I)} \quad \left \| \mathbf{W}(h) -  \mathbf{W}(0)\right\| \leq \omega,  \left \| \mathbf{W}^{(i)}(h) -  \mathbf{W}(0)\right\| \leq \omega, \quad \forall i \in [K]\nonumber, \\ 
      &\textbf{(II)} \quad L(\mathbf{W}(t_c))   \le\left(1-  \Omega\left(\frac{\eta \tau m \phi}{dn^2}\right)\right)^{c} L( \mathbf{W}(0))\nonumber 
 \end{align}
 where $\omega = O\big(\phi^{3/2} n^{-6} L^{-6} \log^{-3/2}(m)\big)$, and $t_c$ is the latest communication round of $h$, which is also $c$th communication round. Then we shall show the above two statements hold for $t+1$.
  \subsubsection{Proof of inductive hypothesis I}
  \paragraph{Step 1: Bounded virtual average iterates.}
  First we prove the first hypothesis for $t+1$: $\left \| \mathbf{W}(t+1) -  \mathbf{W}(0)\right\| \leq \omega $. By the updating rule we know that:
   \begin{align*}
      \left \| \mathbf{W}(t+1) -  \mathbf{W}(0)\right\| &\leq \eta \sum_{j=1}^t\left \|\frac{1}{K}\sum_{i=1}^K \nabla L_i (\mathbf{W}^{(i)}(j))\right\| \nonumber\\
      &\stackrel{\text{\ding{192}}}{\leq} \eta\sum_{j=1}^t \frac{1}{K}\sum_{i=1}^K O(\sqrt{\frac{m}{d}})\sqrt{L_i(\mathbf{W}(j) )}    \nonumber \\
      & \stackrel{\text{\ding{193}}}{\leq} \eta \tau\sum_{j=1}^c \frac{1}{K}\sum_{i=1}^K O(\sqrt{\frac{m}{d}})\sqrt{L_i(\mathbf{W}(t_c) )}\\
      & \leq \eta \tau \sum_{j=1}^c   O(\sqrt{\frac{m}{d}})\sqrt{L(\mathbf{W}(t_c) )},
 \end{align*}
 where we apply the gradient upper bound (Lemma~\ref{lemma:grad bounds}) in \ding{192} and the decreasing nature of local loss (Lemma~\ref{lm: shrink of local loss GD}) in \ding{193}. 
 Now we plug in induction hypothesis \textbf{II} to bound $L(\mathbf{W}(t_c) )$:
    \begin{align*}
      \left \| \mathbf{W}(t+1) -  \mathbf{W}(0)\right\|  
      & \leq \eta \tau \sum_{j=1}^c   O(\sqrt{\frac{m}{d}})\sqrt{ \left(1-  \Omega\left(\frac{\eta \tau m \phi}{dn^2}\right)\right)^{c} L( \mathbf{W}(0))}\\
      & \leq \eta \tau  O(\sqrt{\frac{m}{d}}) \sum_{j=1}^c\left(1-  \Omega\left(\frac{\eta\tau  m \phi}{2dn^2}\right)\right)^{c}\sqrt{ L( \mathbf{W}(0))}\\
      & \leq \eta \tau   O(\sqrt{\frac{m}{d}})   O\left(\frac{2dn^2}{\eta \tau m \phi}\right) \sqrt{ L( \mathbf{W}(0))}\\
      & =     O\left(\frac{2\sqrt{d}n^2}{ \sqrt{m} \phi}\right) \sqrt{ L( \mathbf{W}(0))}.
 \end{align*}
Since we choose $m \geq \frac{d n^{16} L^{12} \log^3 m}{\phi^5}$, it follows that  $ \left \| \mathbf{W}(t+1) -  \mathbf{W}(0)\right\|  \leq \omega$.
 
 \paragraph{Step 2: Bounded local iterates.}
  Then we prove the second hypothesis for $t+1$: $\left \| \mathbf{W}^{(i)}(t+1) - \mathbf{W}(0) \right\| \leq \omega$. By the updating rule we know that:
   \begin{align*} 
      \left \| \mathbf{W}^{(i)}(t+1) -  \mathbf{W}(0)\right\| &\leq \eta \sum_{j=1}^t\left \|  \nabla L_i (\mathbf{W}^{(i)}(j))\right\| \nonumber\\
      &\leq \eta \sum_{j=1}^t   O\left(\sqrt{\frac{m}{d}}\right)\sqrt{L_i(\mathbf{W}(j) )}    \nonumber \\
      & \leq \eta \tau\sum_{j=1}^c  O\left(\sqrt{\frac{m}{d}}\right)\sqrt{L_i(\mathbf{W}(t_c) )}\\
      & \leq \eta \tau \sum_{j=1}^c   O\left(\sqrt{\frac{m}{d}}\right)\sqrt{K L_i(\mathbf{W}(t_c) )},
 \end{align*}
 where we apply the gradient upper bound (Lemma~\ref{lemma:grad bounds}) and the decreasing nature of local loss (Lemma~\ref{lm: shrink of local loss GD}). 
 Now we plug in induction hypothesis \textbf{II} to bound $L(\mathbf{W}(t_c) )$:
    \begin{align*}
      \left \| \mathbf{W}(t+1) -  \mathbf{W}(0)\right\|  
      & \leq \eta \tau \sum_{j=1}^c   O\left(\sqrt{\frac{m}{d}}\right)\sqrt{ \left(1-  \Omega\left(\frac{\eta \tau m \phi}{dn^2}\right)\right)^{c} L_i( \mathbf{W}(0))}\\
      & \leq \eta \tau  O\left(\sqrt{K\frac{m}{d}}\right) \sum_{j=1}^c\left(1-  \Omega\left(\frac{\eta\tau  m \phi}{2dn^2}\right)\right)^{c}\sqrt{ L_i( \mathbf{W}(0))}\\
      & \leq \eta \tau   O\left(\sqrt{K\frac{m}{d}}\right)   O\left(\frac{2dn^2}{\eta \tau m \phi}\right) \sqrt{ L_i( \mathbf{W}(0))}\\
      & =     O\left(\frac{2\sqrt{d}n^2}{ \sqrt{m} \phi}\right) \sqrt{K L_i( \mathbf{W}(0))}.
 \end{align*}
Since we choose $m \geq \frac{K d n^{12} L^{12} \log^3 m}{\phi^5}$, it immediately follows that  $ \left \| \mathbf{W}^{(i)}(t+1) -  \mathbf{W}(0)\right\|  \leq \omega$ as desired.
 
  \subsubsection{Proof of inductive hypothesis II}
  \paragraph{Step 1: One iteration analysis from Semi-smoothness.}
Now we proceed to prove that hypothesis \textbf{II} holds for $t+1$. If $t_c \leq t+1 < t_{c+1}$, then the statement apparently holds for $t_c$. If $t+1\geq t_{c+1}$, we have to examine the upper bound for $L(\mathbf{W}(t_{c+1}))$. The first step is to characterize how global loss changes in one iteration. We use the technique from standard smooth non-convex optimization, but notice that here we only have semi-smooth objective.  According to semi-smoothness (Lemma~\ref{lemma:semi_smooth}) and updating rule:
\begin{align*} 
&L(\mathbf{W}(t_{c+1}))\le L( \mathbf{W}(t_{c})) + \left\langle \nabla L( \mathbf{W}(t_{c})),   \mathbf{W}(t_{c+1})-  \mathbf{W}(t_{c})\right \rangle \notag\\
&\quad+ C'\sqrt{L(\mathbf{W}(t_{c}))}\cdot\frac{\omega^{1/3}L^2\sqrt{m\log(m)}}{\sqrt{d}}\cdot \|\mathbf{W}(t_{c+1})-  \mathbf{W}(t_{c})\|_2 + \frac{C''L^2m}{d}\|\mathbf{W}(t_c)-  \mathbf{W}(t_{c})\|^2_2\\
&\le L( \mathbf{W}(t_{c})) - \left\langle \nabla L( \mathbf{W}(t_{c})),    \eta \tau \frac{1}{\tau K}\sum_{i=1}^K \sum_{t'=t_{c}}^{t_{c+1}-1} \nabla  L_i(\mathbf{W}^{(i)}(t')) \right \rangle \notag\\
&\quad+\eta\tau C'\sqrt{ L(\mathbf{W}(t_{c-1}))}\cdot\frac{\omega^{1/3}L^2\sqrt{m\log(m)}}{\sqrt{d}}\cdot \left\|\frac{1}{\tau K}\sum_{i=1}^K\sum_{t'=t_{c}}^{t_{c+1}-1} \nabla  L_i(\mathbf{W}^{(i)}(t'))\right \| \\
&\quad + \eta^2\frac{C''L^2m}{d}\left\|\frac{1}{K}\sum_{i=1}^K \sum_{t'=t_{c}}^{t_{c+1}-1}\nabla  L_i(\mathbf{W}^{(i)}(t'))\right\|^2 \\
&\stackrel{\text{\ding{192}}}{\leq} L( \mathbf{W}(t_{c})) -  \frac{\eta\tau}{2}\left\| \nabla L( \mathbf{W}(t_{c})) \right\|^2_{\mathrm{F}} -\frac{\eta\tau}{2} \left\| \frac{1}{\tau K}\sum_{i=1}^K\sum_{t'=t_{c}}^{t_{c+1}-1} \nabla  L_i(\mathbf{W}^{(i)}(t)) \right \|^2_{\mathrm{F}} \\
&\quad +\frac{\eta\tau}{2} \left\| \nabla L( \mathbf{W}(t_c))-\frac{1}{\tau K}\sum_{i=1}^K \sum_{t'=t_{c}}^{t_{c+1}-1}\nabla  L_i(\mathbf{W}^{(i)}(t')) \right \|^2_{\mathrm{F}}  \\
&\quad+\eta \tau C'\sqrt{ L(\mathbf{W}(t_c))} \cdot\frac{\omega^{1/3}L^2\sqrt{m\log(m)}}{2\sqrt{d}}\\
&\quad \times \left( \left\|  \nabla  L(\mathbf{W}(t_c))\right \|_{\mathrm{F}}+\left\|\frac{1}{\tau K}\sum_{i=1}^K\sum_{t'=t_{c}}^{t_{c+1}-1} \nabla  L_i(\mathbf{W}^{(i)}(t'))-  \nabla  L(\mathbf{W}(t_c))\right \|_{\mathrm{F}}\right) \\
&\quad+ \eta^2 \frac{C''L^2m}{d}\left\|\frac{1}{ K}\sum_{i=1}^K\sum_{t'=t_{c}}^{t_{c+1}-1} \nabla  L_i(\mathbf{W}^{(i)}(t'))\right\|^2_{\mathrm{F}} \\
&\le L( \mathbf{W}(t_c)) -  \frac{\eta \tau}{2}\left\| \nabla L( \mathbf{W}(t_c)) \right\|^2_{\mathrm{F}} -\left(\frac{\eta}{2\tau} -  \frac{\eta^2 C''L^2m}{d}\right) \left\| \frac{1}{K}\sum_{i=1}^K \sum_{t'=t_{c}}^{t_{c+1}-1} \nabla  L_i(\mathbf{W}^{(i)}(t')) \right \|^2_{\mathrm{F}}\\
& \quad + \frac{\eta }{2} \frac{1}{ K}\sum_{i=1}^K \sum_{t'=t_{c}}^{t_{c+1}-1}\left\| \nabla L_i( \mathbf{W}(t_c))-\nabla  L_i(\mathbf{W}^{(i)}(t')) \right \|^2_{\mathrm{F}}  +\eta C'\sqrt{ L(\mathbf{W}(t))} \cdot\frac{\omega^{1/3}L^2\sqrt{m\log(m)}}{2\sqrt{d}}\left\|  \nabla  L(\mathbf{W}(t_c))\right \|_{\mathrm{F}}  \\
&\quad+\eta \tau C' \cdot\frac{\omega^{1/3}L^2\sqrt{m\log(m)}}{2\sqrt{d}}\left(  \left\|\nabla L( \mathbf{W}(t_c))-\frac{1}{\tau K}\sum_{i=1}^K \sum_{t'=t_{c}}^{t_{c+1}-1}\nabla  L_i(\mathbf{W}^{(i)}(t'))  \right \|_{\mathrm{F}} \sqrt{L( \mathbf{W}(t_c))}\right),   
\end{align*}
where in \ding{192} we use the identity $\langle \bm{a},\bm{b} \rangle =  \frac{1}{2}\|\bm{a} \|^2+\frac{1}{2}\| \bm{b}\|^2 - \frac{1}{2}\|\bm{a}-\bm{b}\|^2$. 
We plug in the semi gradient Lipschitzness from Lemma~\ref{lm: gradient gap}   and gradient bound from Lemma~\ref{lemma:grad bounds} in last inequality, and use the fact that $\frac{\eta}{2\tau} -  \frac{\eta^2 C''L^2m}{d}\geq 0$ to get:

\begin{align*} 
L(\mathbf{W}(t_{c+1}))  
&\le L( \mathbf{W}(t_{c})) -   \Omega \left(\frac{\eta \tau m\phi}{dn^2}  \right) L(\mathbf{W}_{t_{c}} ) \\
& \quad + \frac{\eta \tau }{2}   \left(O\left( \frac{m L^4}{d} \left\|\mathbf{W}^{(i)}(t') -  \mathbf{W}(t_{c})\right\|_2^2\right)+O\left(\frac{\omega^{2/3} L^5 {m\log m }}{d}+ \frac{\omega^2 L^{6 } {m \log m}}{d}\right) L(\mathbf{W}(t_{c}))\right) \\
& \quad +\eta \tau C' \cdot\frac{\omega^{1/3}L^2 {m\log(m)}}{2 {d}} L(\mathbf{W}({t_{c}}) ) \\
&\quad+\eta \tau C' \cdot\frac{\omega^{1/3}L^2\sqrt{m\log(m)}}{2\sqrt{d}}\\
& \quad \times \left( \sqrt{O\left( \frac{m L^4}{d} \left\|\mathbf{W}^{(i)}(t') -  \mathbf{W}(t_{c})\right\|_2^2\right)+O\left(\frac{\omega^{2/3} L^5 {m\log m }}{d}+ \frac{\omega^2 L^{6 } {m \log m}}{d}\right) L(\mathbf{W}(t_{c})) }L( \mathbf{W}(t_{c}) \right),   
\end{align*}
Choosing $\omega = \frac{ \phi^{3/2} }{C_\omega n^{6}L^{6}\log(m)^{3/2}}$ where $C_\omega$ is some large constant and plugging in local model deviation bound from Lemma~\ref{lemma: weight gap GD},   to  get the main recursion relation as follows:
\begin{align} 
L(\mathbf{W}(t_{c+1}))  
&\le \left(1-  \Omega\left(\frac{\eta \tau  m \phi}{dn^2}\right)\right)L( \mathbf{W}(t_{c}))  .\label{eq: proof_GD_1}
\end{align}
Unrolling the recursion and plugging in $\eta = O(\frac{d n ^2}{\tau m \phi})$ will conclude the proof:
\begin{align*} 
&L(\mathbf{W}(T))  \le \exp (-R) L( \mathbf{W}(0)) = \epsilon\\
\Longleftrightarrow &R = O\left( \log \frac{1}{\epsilon}\right).
\end{align*}

\section{Proof of Theorem~\ref{theorem: local SGD} (Local SGD)}\label{app: SGD}
In this section  we will present the proof of convergence rate of Local SGD (Theorem~\ref{theorem: local SGD}). Before that, let us first introduce some useful lemmas. 
\subsection{Proof of Technical Lemma}
The following lemma establishes the boundedness of the stochastic gradient.
\begin{lemma}[\sffamily{Bounded stochastic gradient}]\label{lemma: bounded variance}
For Local SGD, the following statement holds true for stochastic gradient at any iteration $t$:
\begin{align*}
    \mathbb{E}_{S_t}\left[ \left \| \frac{1}{K}\sum_{i=1}^K \mathbf{G}_i^{(t)}  \right\|^2 \right] \leq  O\left(\frac{mL (\mathbf{W}(t_c))}{dK}\right),     \left \| \frac{1}{K}\sum_{i=1}^K \mathbf{G}_i^{(t)}  \right\|^2   \leq  O\left(\frac{mnL (\mathbf{W}(t_c))}{d}\right),
\end{align*}
where $S_t = \{(\tilde{\bm{x}}_i, \tilde{y}_i)\}_{i=1}^K$ are the set of randomly sampled data to compute $\frac{1}{K}\sum_{i=1}^K \mathbf{G}_i^{(t)}$, $K$ is the number of devices, $\tau$ is the number of local updates between two consecutive rounds of synchronization, $n$ is the size of each local data shard, $d$ is the dimension of input data, and $m$ is the number of neurons in hidden layer.
\begin{proof}
Observe the following facts:
\begin{align*}
    \mathbb{E}_{S_t}\left[ \left \|\frac{1}{K}\sum_{i=1}^K \mathbf{G}_i^{(t)} \right\|^2 \right]  
    &= \frac{1}{K^2}\sum_{i=1}^K\mathbb{E}\left[ \left \| \nabla \ell (\mathbf{W}^{(i)}(t); \tilde{\bm{x}}_i, \tilde{y}_i)  \right\|^2 \right]\\
   & = \frac{1}{K^2}\sum_{i=1}^K\frac{1}{n}\sum_{( \bm{x}_j , y_j)\in S_i} \left \| \nabla \ell (\mathbf{W}^{(i)}(t); \bm{x}_j , y_j)   \right\|^2    \\ 
   & \leq \frac{1}{K^2}\sum_{i=1}^K O(m/d) L_i (\mathbf{W}^{(i)}(t)),
\end{align*}
where we plug in the gradiet upper bound from Lemma~\ref{lemma:grad bounds}.
According to the shrinkage of local loss (Lemma~\ref{lm: shrink of local loss SGD}), we can conclude that:
\begin{align*}
    &\mathbb{E}_{S_t}\left[ \left \|\frac{1}{K}\sum_{i=1}^K \mathbf{G}_i^{(t)} \right\|^2 \right]   \leq  O\left(\frac{mL (\mathbf{W}(t_c))}{dK}\right) .
\end{align*}
Now we switch to prove the second statement by observing that:
\begin{align*}
    \left \|\frac{1}{K}\sum_{i=1}^K \mathbf{G}_i^{(t)} \right\|^2  
    &= \frac{1}{K}\sum_{i=1}^K\left \| \nabla \ell (\mathbf{W}^{(i)}(t); \tilde{\bm{x}}_i, \tilde{y}_i)  \right\|^2 \\
   & \leq \frac{1}{K}\sum_{i=1}^K O(mn/d) L_i (\mathbf{W}^{(i)}(t))\\
   & \leq \frac{1}{K}\sum_{i=1}^K O(mn/d) L_i (\mathbf{W}^{(i)}(t_c)),
\end{align*}
which completes the proof.
\end{proof}
\end{lemma}

The next lemma is similar to Lemma~\ref{lemma: weight gap GD}, but it characterizes the local model deviation under stochastic setting. Hence, it will be inevitably looser than the deterministic version (Lemma~\ref{lemma: weight gap GD}).  
\begin{lemma}\label{lemma: weight gap SGD}
For Local SGD, let $t_c$ denote  the latest communication stage of before iteration $t$. If the condition that $L_i(\mathbf{W}^{(i)}(t) ) \leq L_i(\mathbf{W}(t_c) )$ for any $t_c \leq t\leq t_c + \tau -1$ holds, then the following statement holds true for $t_c \leq t \leq t_c + \tau-1$: 
\begin{align*}
 \left\| \mathbf{W}^{(i)}(t) -  \mathbf{W}(t_c) \right\|^2_{\mathrm{F}}  
    & \leq     O\left((\eta^2 \tau^2   + \eta^2 \tau)\frac{mn}{d}\right) L_i(\mathbf{W}(t_c) ), 
\end{align*}
where $K$ is the number of devices, $\tau$ is the number of local updates between two consecutive rounds of synchronization, $n$ is the size of each local data shard, and $m$ is the number of neurons in hidden layer.
\begin{proof}
According to updating rule:
\begin{align*}
    \mathbb{E}\left\| \mathbf{W}^{(i)}(t+1) -  \mathbf{W}(t_c) \right\|^2_{\mathrm{F}} &=  \mathbb{E}\left\| \mathbf{W}^{(i)}(t) - \eta  \mathbf{G}^{(t)}_i - \mathbf{W}(t_c) \right\|^2_{\mathrm{F}}  \\
    & =  \mathbb{E}\left\| \mathbf{W}^{(i)}(t) - \mathbf{W}(t_c) \right\|^2_{\mathrm{F}} - 2\eta  \mathbb{E} \left \langle \nabla L_i(\mathbf{W}^{(i)}(t)), \mathbf{W}^{(i)}(t) - \mathbf{W}(t_c) \right\rangle\\
    &\quad  + \eta^2  \mathbb{E}\|\mathbf{G}^{(t)}_i\|^2_\mathrm{F}\\
    & =  \mathbb{E}\left\| \mathbf{W}^{(i)}(t) - \mathbf{W}(t_c) \right\|^2_{\mathrm{F}} + 2\eta \left \langle \nabla L_i(\mathbf{W}^{(i)}(t)),  \eta \sum_{t' = t_c}^{t-1} \nabla L_i(\mathbf{W}^{(i)}(t)) \right\rangle\\
    &\quad  + \eta^2 \mathbb{E} \|\mathbf{G}^{(t)}_i\|^2_\mathrm{F}\\
     & =  \mathbb{E}\left\| \mathbf{W}^{(i)}(t) - \mathbf{W}(t_c) \right\|^2_{\mathrm{F}} + 2\eta^2 (t-t_c) \left \langle \nabla L_i(\mathbf{W}^{(i)}(t)),   \frac{1}{(t-t_c)} \sum_{t' = t_c}^{t-1} \nabla L_i(\mathbf{W}^{(i)}(t)) \right\rangle\\
    &\quad  + \eta^2  \mathbb{E}\|\mathbf{G}^{(t)}_i\|^2_\mathrm{F}.
\end{align*}
Applying the identity $\langle \bm{a}, \bm{b} \rangle = \frac{1}{2}\|\bm{a}\|^2+\frac{1}{2}\|\bm{b}\|^2 - \frac{1}{2}\|\bm{a}-\bm{b}\|^2$ on the cross term we have:
 \begin{align*}
     \mathbb{E}\left\| \mathbf{W}^{(i)}(t+1) -  \mathbf{W}(t_c) \right\|^2_{\mathrm{F}}   
     & \leq  \mathbb{E}\left\| \mathbf{W}^{(i)}(t) - \mathbf{W}(t_c) \right\|^2_{\mathrm{F}} \\
     & \quad +  \eta^2 (t-t_c) \left( \mathbb{E}\left \| \nabla L_i(\mathbf{W}^{(i)}(t))\right\|^2+ \left \|   \frac{1}{(t-t_c)} \sum_{t' = t_c}^{t-1} \nabla L_i(\mathbf{W}^{(i)}(t)) \right\|^2 \right)\\
    &\quad  + \eta^2 \frac{1}{n} \sum_{( \bm{x}_j , y_j)\in S_i} \left \| \nabla \ell (\mathbf{W}^{(i)}(t); \bm{x}_j , y_j)   \right\|^2.
\end{align*}
Plugging the gradient upper bound from Lemma~\ref{lemma:grad bounds} yields:
 \begin{align*}
   \mathbb{E} \left\| \mathbf{W}^{(i)}(t+1) -  \mathbf{W}(t_c) \right\|^2_{\mathrm{F}}  
    & \leq \mathbb{E} \left\| \mathbf{W}^{(i)}(t) - \mathbf{W}(t_c) \right\|^2_{\mathrm{F}}\\
    &\quad +  \eta^2 (t-t_c) \left(O\left(\frac{m}{d}\right) L_i(\mathbf{W}^{(i)}(t) ) + \frac{1}{t-t_c} \sum_{t' = t_c}^{t-1}O\left(\frac{m}{d}\right) L_i(\mathbf{W}^{(i)}(t') )  \right)\\
    &\quad  + \eta^2 O\left(\frac{m}{d}\right) L_i(\mathbf{W}^{(i)}(t) ). 
\end{align*}
Since we assume $L_i(\mathbf{W}^{(i)}(t) ) \leq L_i(\mathbf{W}(t_c) )$ for any $t_c \leq t\leq t_c + \tau -1$, we have:
 \begin{align*}
    \mathbb{E}\left\| \mathbf{W}^{(i)}(t+1) -  \mathbf{W}(t_c) \right\|^2_{\mathrm{F}} 
    & \leq \mathbb{E}\left\| \mathbf{W}^{(i)}(t) - \mathbf{W}(t_c) \right\|^2_{\mathrm{F}} +  \eta^2 \tau \left(O\big(\frac{m }{d}\big) L_i(\mathbf{W}^{(i)}(t_c) ) +  O\big(\frac{m }{d}\big) L_i(\mathbf{W}^{(i)}(t_c) )  \right)\\
    &\quad  + \eta^2 O\big(\frac{m }{d}\big) L_i(\mathbf{W}(t_c) ).  
\end{align*}
Do the telescoping  sum from $t+1$ to $t_c$ will conclude the proof:
 \begin{align*}
    \mathbb{E}\left\| \mathbf{W}^{(i)}(t+1) -  \mathbf{W}(t_c) \right\|^2_{\mathrm{F}}  
    & \leq     O\left((\eta^2 \tau^2   + \eta^2 \tau)\frac{m }{d}\right) L_i(\mathbf{W}(t_c) ). 
\end{align*} 
\end{proof} 
\end{lemma}

The next lemma will reveal the boundedness of objective in stochastic setting. The slight difference to the dynamic of objective in deterministic setting (Lemma~\ref{lm: shrink of local loss GD}) is that, we show the objective is strictly decreasing in Local GD, but here we only derive a small upper bound of it: $L_i(\mathbf{W}^{(i)}(t))  \leq   O(1)\cdot L_i(\mathbf{W}(t_c))$, with high probability. Even though it is not a strictly decreasing loss, it is enough to enable us to prove linear convergence of objective.

 \begin{lemma}\label{lm: shrink of local loss SGD}
 For Local SGD, at any iteration $t$ in between two communication rounds: $t_c \leq t\leq t_c + \tau -1$ and $i \in [K]$, if $\mathbf{W}^{(i)}(t) \in \mathcal{B}(\bm{W}(0),\omega)$, then with probability at least $1-\exp\left(-\Omega(m\omega^{2/3})\right)$, the following statement holds true:
 \begin{align*}
   L_i(\mathbf{W}^{(i)}(t))  \leq   O(1)\cdot L_i(\mathbf{W}(t_c)) .
 \end{align*} 
 \begin{proof}
%  \begin{align*} 
% &\mathbb{E}[L_i(\mathbf{W}^{(i)}(t))]\le L_i( \mathbf{W}^{(i)}(t-1)) + \mathbb{E}\left[\left\langle \nabla L_i( \mathbf{W}^{(i)}(t-1)),   \mathbf{W}^{(i)}(t)-  \mathbf{W}^{(i)}(t-1)\right \rangle \right]\notag\\
% &\quad+ C'\sqrt{L_i(\mathbf{W}^{(i)}(t))}\cdot\frac{\omega^{1/3}L^2\sqrt{m\log(m)}}{\sqrt{d}}\cdot \mathbb{E}\|\mathbf{W}^{(i)}(t)-  \mathbf{W}^{(i)}(t-1)\|_2 + \frac{C''L^2m}{d}\mathbb{E}\|\mathbf{W}^{(i)}(t)-  \mathbf{W}^{(i)}(t-1)\|^2_2\\
% &\le L( \mathbf{W}^{(i)}(t-1)) -\eta  \left\langle \nabla L( \mathbf{W}^{(i)}(t-1)),     \nabla  L_i(\mathbf{W}^{(i)}(t-1))  \right \rangle \notag\\
% &\quad+\eta  C'\sqrt{ L_i(\mathbf{W}^{(i)}(t-1))}\cdot\frac{\omega^{1/3}L^2\sqrt{m\log(m)}}{\sqrt{d}}\cdot \mathbb{E}\left\|  \mathbf{G}^{(i)}(t)\right \|_2   + \eta^2\frac{C''L^2m}{d}\mathbb{E}\left\|  \mathbf{G}^{(i)}(t)\right\|^2_2  
% \end{align*}
% Using Lemma~\ref{lemma: bounded variance} to bound $\mathbb{E}\left\|  \mathbf{G}^{(i)}(t)\right\|^2_2$ yields:
% \begin{align} 
% &\mathbb{E}[L_i(\mathbf{W}^{(i)}(t))]  \le L_i( \mathbf{W}(t-1)) -\eta \left\|  \nabla L_i( \mathbf{W}^{(i)}(t-1)) \right\|^2_\mathrm{F}   \notag\\
% &\quad+\eta  C'\sqrt{ L(\mathbf{W}^{(i)}(t-1))}\cdot\frac{\omega^{1/3}L^2\sqrt{m\log(m)}}{\sqrt{d}}\cdot O\left(\sqrt{\frac{mL (\mathbf{W}(t_c))}{dK}}\right) + \eta^2\frac{C''L^2m}{d}O\left(\frac{mL (\mathbf{W}(t_c))}{dK}\right)  \nonumber\\
% & \le \left(1- \Omega \left(\frac{\eta m \phi}{d n^2}\right) \right)L_i(\mathbf{W}(t-1)).\label{eq: SGD shrinkage loss 1}
% \end{align}

We examine the absolute value bound for $L_i(\mathbf{W}^{(i)}(t-1))$:
 \begin{align} 
L_i(\mathbf{W}^{(i)}(t)) &\le L_i( \mathbf{W}^{(i)}(t-1)) + \eta \left\| \nabla L_i( \mathbf{W}(t-1))\right \|\left\| \mathbf{G}^{(i)}(t)  \right \| \notag\\ 
&\quad+\eta  C'\sqrt{ L_i(\mathbf{W}^{(i)}(t-1))}\cdot\frac{\omega^{1/3}L^2\sqrt{m\log(m)}}{\sqrt{d}}\cdot  \left\|  \mathbf{G}^{(i)}(t)\right \|_2   + \eta^2\frac{C''L^2m}{d} \left\|  \mathbf{G}^{(i)}(t)\right\|^2_2 \nonumber \\
&\le  L_i( \mathbf{W}^{(i)}(t-1)) +   \frac{\eta m\sqrt{n}}{d} L_i(\mathbf{W}^{(i)}(t-1)) \notag\\ 
&\quad+\eta  C'\frac{\omega^{1/3}L^2m\sqrt{mn \log(m)}}{d\sqrt{d}} L_i( \mathbf{W}^{(i)}(t-1))  + \eta^2\frac{C''L^2m^2 n}{d^2} L_i( \mathbf{W}(t-1)) \nonumber  \\
& \le \left(1+ O \left(\frac{\eta m \sqrt{n}}{d }\right) \right) L_i( \mathbf{W}^{(i)}(t-1)) \nonumber\\
& \leq \left(1+ O \left(\frac{ \phi}{m\tau n^{2.5} \log^2 m  }   \right) \right)^\tau L_i( \mathbf{W}(t_c))\nonumber\\
& \leq \exp \left(\frac{ \phi}{m  n^{2.5} \log^2 m  } \right)  L_i( \mathbf{W}(t_c)) \leq  O(1)\cdot L_i( \mathbf{W}(t_c))\nonumber.
\end{align}
 \end{proof}
 
 \end{lemma}

 \subsection{Proof of Theorem~\ref{theorem: local SGD}}
 With the above lemmas in hand, we can finally proceed to the proof of Theorem~\ref{theorem: local SGD}.
 We prove Theorem~\ref{theorem: local SGD} by induction. Assume the following induction hypotheses hold for all $h \leq t$, with probability at least $1-e^{-\Omega( (\log m)^2)}$:
 \begin{align}
      &\textbf{(I)} \quad \left \| \mathbf{W}(h) -  \mathbf{W}(0)\right\| \leq \omega, \left \| \mathbf{W}^{(i)}(h) -  \mathbf{W}(0)\right\| \leq \omega, \forall i \in [K] \nonumber, \\
      &\textbf{(II)} \quad L(\mathbf{W}(t_c))   \le  {n\log^2 m}\cdot e^{-R/R_0} \nonumber,
 \end{align}
 where $\omega = O\big(\phi^{3/2}n^{-6}L^{-6} \log^{-3/2}(m)\big)$, $t_c$ is the latest communication round of $h$, also the $c$th communication round, and $R_0 = \frac{ n^5\log^2m}{\phi^2}$. Then, we need to show that these two statements hold for $t+1$.
  \subsubsection{Proof of inductive hypothesis I}
  \paragraph{Step 1: Bounded virtual average iterates.}
  Now we prove the first hypothesis for $t+1$: $\left \| \mathbf{W}(t+1) -  \mathbf{W}(0)\right\| \leq \omega $. By the updating rule we know that:
  \begin{align*}
      \left \| \mathbf{W}(t+1) -  \mathbf{W}(0)\right\| &\leq \eta \sum_{j=1}^t\left \|\frac{1}{K}\sum_{i=1}^K \mathbf{G}^{(i)}(j)\right\| \nonumber\\
      &\leq \eta \tau \sum_{c'=1}^c   O\left(\sqrt{\frac{mn}{d} L(\mathbf{W}(t_c')}\right)     \nonumber \\
      & \leq \eta \tau \sum_{c'=1}^c   O\left(\sqrt{ \frac{mn}{d}  {n\log^2 m}e^{-c'/R_0}}\right)     \nonumber \\ 
       & \leq \eta \tau  \sqrt{ \frac{mn}{d}  {n\log^2 m} } \left(1+\frac{1}{e^{1/(2R_0)}-1}\right)\nonumber\\
       & \leq \eta \tau  \sqrt{ \frac{mn}{d}  {n\log^2 m} } \left(1+2R_0\right)\nonumber\\
       & \leq \frac{\sqrt{d} \log m n^3}{\sqrt{m}}.
 \end{align*}
 Since we choose $m \geq \frac{n^{18}L^{12}d \log^5 m}{\phi^3}$, we conclude that $\left \| \mathbf{W}(t+1) -  \mathbf{W}(0)\right\| \leq \omega$.
 
 \paragraph{Step 2: Bounded local iterates.}
  Then we prove the second hypothesis for $t+1$: $\left \| \mathbf{W}^{(i)}(t+1) - \mathbf{W}(0) \right\| \leq \omega$. By the updating rule we know that:
   \begin{align*} 
      \left \| \mathbf{W}^{(i)}(t+1) -  \mathbf{W}(0)\right\| &\leq \eta \sum_{j=1}^t\left \| \mathbf{G}^{(i)} (j)\right\| \nonumber\\
      &\leq \eta \sum_{j=1}^t   O\left(\sqrt{\frac{mn}{d}}\right)\sqrt{L_i(\mathbf{W}(j) )}    \nonumber \\
      & \leq \eta \tau\sum_{j=1}^c  O\left(\sqrt{\frac{m}{d}}\right)\sqrt{KL(\mathbf{W}(t_c) )}\\
      & \leq  \frac{\sqrt{dK} \log m n^3}{\sqrt{m}},
 \end{align*}
 where we apply the gradient upper bound (Lemma~\ref{lemma:grad bounds}) and the decreasing nature of local loss (Lemma~\ref{lm: shrink of local loss SGD}).  
Since we choose $m \geq \frac{K d n^{18} L^{12} \log^3 m}{\phi^5}$, we know that $ \left \| \mathbf{W}^{(i)}(t+1) -  \mathbf{W}(0)\right\|  \leq \omega$.

\subsubsection{Proof of inductive hypothesis II}
Now we proceed to prove that hypothesis \textbf{II} holds for $t+1$.  If $t_c \leq t+1 < t_{c+1}$, then the statement apparently holds for $t_c$. If $t+1\geq t_{c+1}$, we have to examine the upper bound for $L(\mathbf{W}(t_{c+1}))$. The first step is to characterize how global loss changes in one iteration. We use the technique from standard smooth non-convex optimization, but notice that here we only have semi-smooth objective.  According to semi-smoothness (Lemma~\ref{lemma:semi_smooth}) and updating rule:
\begin{align*} 
&\mathbb{E}[L(\mathbf{W}(t_{c+1}))]\le L( \mathbf{W}(t_{c})) + \mathbb{E}\left[\left\langle \nabla L( \mathbf{W}(t_{c})),   \mathbf{W}(t_{c+1})-  \mathbf{W}(t_{c})\right \rangle \right]\notag\\
&\quad+ C'\sqrt{L(\mathbf{W}(t_c))}\cdot\frac{\omega^{1/3}L^2\sqrt{m\log(m)}}{\sqrt{d}}\cdot \mathbb{E}\|\mathbf{W}(t_{c+1})-  \mathbf{W}(t_{c})\|_2 + \frac{C''L^2m}{d}\mathbb{E}\|\mathbf{W}(t_{c+1})-  \mathbf{W}(t_{c})\|^2_2\\
&\le L( \mathbf{W}(t_{c})) - \left\langle \nabla L( \mathbf{W}(t_{c})),    \eta \tau \frac{1}{\tau K}\sum_{i=1}^K \sum_{t'=t_{c}}^{t_{c+1}-1} \nabla  L_i(\mathbf{W}^{(i)}(t'))  \right \rangle \notag\\
&\quad+\eta\tau C'\sqrt{ L(\mathbf{W}(t_{c}))}\cdot\frac{\omega^{1/3}L^2\sqrt{m\log(m)}}{\sqrt{d}}\cdot \mathbb{E}\left\|\frac{1}{\tau K}\sum_{i=1}^K\sum_{t'=t_{c}}^{t_{c+1}-1} \mathbf{G}^{(i)}(t')\right \|_2 \\
&\quad + \eta^2\frac{C''L^2m}{d}\mathbb{E}\left\|\frac{1}{\tau K}\sum_{i=1}^K\sum_{t'=t_{c}}^{t_{c+1}-1} \mathbf{G}^{(i)}(t')\right\|^2_2 \\
&\le L( \mathbf{W}(t_{c})) - \left\langle \nabla L( \mathbf{W}(t_{c})),    \eta \tau \frac{1}{\tau K}\sum_{i=1}^K \sum_{t'=t_{c}}^{t_{c+1}-1} \nabla  L_i(\mathbf{W}^{(i)}(t'))  \right \rangle \notag\\
&\quad+\eta\tau C'\sqrt{ L(\mathbf{W}(t_{c}))}\cdot\frac{\omega^{1/3}L^2\sqrt{m\log(m)}}{\sqrt{d}}\cdot  O\left(\sqrt{\frac{mL (\mathbf{W}(t_{c}))}{dK}}\right) \\
&\quad + \eta^2\frac{C''L^2m}{d}  O\left(\frac{mL (\mathbf{W}(t_{c}))}{dK}\right) \\
&\stackrel{\text{\ding{192}}}{\leq} L( \mathbf{W}(t_{c})) -  \frac{\eta\tau}{2}\left\| \nabla L( \mathbf{W}(t_{c})) \right\|^2_{\mathrm{F}} +\eta \tau C'\sqrt{ L(\mathbf{W}(t_{c}))} \cdot\frac{\omega^{1/3}L^2\sqrt{m\log(m)}}{2\sqrt{d}}O\left(\sqrt{\frac{mL (\mathbf{W}(t_{c}))}{dK}}\right) \\
&\quad+ \eta^2 \frac{C''L^2m}{d}O\left(\frac{mL (\mathbf{W}(t_{c}))}{dK}\right)+\frac{\eta }{2} \frac{1}{ K}\sum_{i=1}^K\sum_{t'=t_{c}}^{t_{c+1}-1}\left\| \nabla L_i( \mathbf{W}(t_{c}))-\nabla  L_i(\mathbf{W}^{(i)}(t')) \right \|^2_{\mathrm{F}},
\end{align*}
where in \ding{192} we use the identity $\langle \bm{a},\bm{b} \rangle =  \frac{1}{2}\|\bm{a} \|^2+\frac{1}{2}\| \bm{b}\|^2 - \frac{1}{2}\|\bm{a}-\bm{b}\|^2$. 
We plug in the semi gradient Lipschitzness from Lemma~\ref{lm: gradient gap}   and gradient bound from Lemma~\ref{lemma:grad bounds} in last inequality to get: 
\begin{align*} 
\mathbb{E}[L(\mathbf{W}(t_{c+1}))] 
&\leq L( \mathbf{W}(t_{c})) -  \frac{\eta\tau}{2}\left\| \nabla L( \mathbf{W}(t_{c})) \right\|^2_{\mathrm{F}} \\
&\quad +\eta \tau C'\sqrt{ L(\mathbf{W}(t_{c}))} \cdot\frac{\omega^{1/3}L^2\sqrt{m\log(m)}}{2\sqrt{d}}O\left(\sqrt{\frac{mL (\mathbf{W}(t_{c}))}{dK}}\right) \\
&\quad+ \eta^2 \frac{C''L^2m}{d}O\left(\frac{mL (\mathbf{W}(t_{c}))}{dK}\right)\\
&\quad +\frac{\eta }{2} \frac{1}{ K}\sum_{i=1}^K \sum_{t'=t_{c}}^{t_{c+1}-1}\left(O\left( \frac{m L^4}{d} \left\|\mathbf{W}^{(i)}(t') -  \mathbf{W}(t_{c})\right\|_2^2\right)\right.\\
&\qquad \qquad \qquad \qquad \qquad \left. +O\left(\frac{\omega^{2/3} L^5 {m\log m }}{d}+ \frac{\omega^2 L^{6 } {m \log m}}{d}\right) L(\mathbf{W}(t_{c}))\right),
\end{align*}
Choosing $\omega = \frac{ \phi^{3/2} }{C_\omega n^{6}L^{6}\log(m)^{3/2}}$ where $C_\omega$ is some large constant and plugging in local model deviation bound from Lemma~\ref{lemma: weight gap SGD},  to  get the main recursion relation as follows:
\begin{align} 
\mathbb{E}[L(\mathbf{W}(t_{c+1}))]  
&\le \left(1-  \Omega\left(\frac{\eta \tau m \phi}{dn^2}\right)\right)L( \mathbf{W}(t_{c}))  .
  \label{eq: proof_SGD_1}
\end{align}
 
Also by semi smoothness, we have:
\begin{align}
&L(\mathbf{W}(t_{c+1}))  
\le L( \mathbf{W}(t_{c})) + 2\left\| \nabla L(\mathbf{W}(t_{c})) \right\|_\mathrm{F}\left\| \frac{1}{K}\sum_{i=1}^K\sum_{t'=t_{c}}^{t_{c+1}-1} \mathbf{G}^{(i)}(t')\right\|_\mathrm{F} \nonumber\\
& \quad + C'\eta \sqrt{L(\mathbf{W}(t_c))}\cdot\frac{\omega^{1/3}L^2\sqrt{m\log(m)}}{\sqrt{d}}\cdot \left\|\frac{1}{K}\sum_{i=1}^K\sum_{t'=t_{c}}^{t_{c+1}-1} \mathbf{G}^{(i)}(t')\right\|_2 + \frac{C''\eta^2 L^2m}{d}\mathbb{E}\left\|\frac{1}{K}\sum_{i=1}^K\sum_{t'=t_{c}}^{t_{c+1}-1} \mathbf{G}^{(i)}(t')\right\|^2_2\nonumber\\ 
& \le L( \mathbf{W}(t_{c})) + 2 \sqrt{\frac{m}{d}} \sqrt{L(\mathbf{W}(t_{c}))}  \tau \sqrt{\frac{mn}{d}} \sqrt{L(\mathbf{W}(t_{c}))} \nonumber\\
& \quad + C'\eta \sqrt{L(\mathbf{W}(t_c))}\cdot\frac{\omega^{1/3}L^2\sqrt{m\log(m)}}{\sqrt{d}}\cdot \tau \sqrt{\frac{mn}{d}} \sqrt{L(\mathbf{W}(t_{c}))} + \frac{C''\eta^2 \tau^2 L^2m^2}{d^2 K}   L(\mathbf{W}(t_{c})) \nonumber\\
& \leq\left[ 1+ O\left(\frac{\eta m\tau \sqrt{n}}{d}\right)\right]L( \mathbf{W}(t_{c})) \label{eq: proof_SGD_2},
\end{align}

Taking log on the both sides of (\ref{eq: proof_SGD_1}) and (\ref{eq: proof_SGD_2})  yields:
 \begin{align*} 
\log [L(\mathbf{W}(t_{c+1}))]   & \le \log [L( \mathbf{W}(t_{c}))] +   O \left(\frac{\eta m \tau \sqrt{n}}{d}\right) ,\\ 
\mathbb{E}[\log [L(\mathbf{W}(t_{c+1}))]]   &\le \log\mathbb{E}[  L(\mathbf{W}(t_{c+1}))] \le \log [L( \mathbf{W}(t_{c}))] + \log  \left(1- \Omega \left(\frac{\eta m \tau \phi}{d n^2}\right) \right) \\
& \le \log [L( \mathbf{W}(0))] - c  \Omega \left(\frac{\eta m \tau \phi}{d n^2}\right),
\end{align*}

So we can apply martingale concentration inequality. With probability at least $1- e^{-\Omega(\log^2 m)}$
\begin{align*}
    \log [L(\mathbf{W}(t_{c+1}))] &\leq \mathbb{E}[\log [L(\mathbf{W}(t_{c})]]  + \sqrt{c} O \left(\frac{\eta m \tau \sqrt{n}}{d }\right) \log m \\
    & \leq \log [L( \mathbf{W}(0))] - c\Omega \left(\frac{\eta m \tau\phi}{d n^2}\right) + \sqrt{c} O \left(\frac{\eta m \tau \sqrt{n}}{d }\right) \log m \\
    & \leq \log [L( \mathbf{W}(0))] - \left(\sqrt{c}\Omega \left(\sqrt{\frac{\eta m \tau\phi}{d n^2}}\right) - \sqrt{\frac{d n^2}{\eta m \tau \phi}} O \left(\frac{\eta m \tau \sqrt{n}}{d }\right) \log m\right)^2  + O \left(\frac{\eta  m \tau n^3}{d\phi } \log^2m\right),
\end{align*}
where in the last inequality we use the fact that $2a\sqrt{c} - b^2 c = -(b\sqrt{c}-a/b)^2+a^2/b^2$. Plugging that $\eta = \frac{d \phi}{m\tau n^3 \log^2 m}$ yields:
\begin{align*}
    \log [L(\mathbf{W}(t_{c+1}))]  
    & \leq \log [L( \mathbf{W}(0))] - \left(\sqrt{c}\Omega \left(\sqrt{\frac{\eta m \tau \phi}{d n^2}}\right) - \sqrt{\frac{d n^2}{\eta m \tau\phi}} O \left(\frac{\eta m \tau \sqrt{n}}{d }\right) \log m\right)^2 + O \left(1\right)\\
    & \leq \log [L( \mathbf{W}(0))] - \mathbf{1}\left[ c\geq \Theta\left(\frac{n^5 \log^2 m}{\phi^2}\right)\right] \Omega \left( {\frac{\eta m \tau \phi}{d n^2}}c\right) + O \left(1\right)\\
     & \leq \log [L( \mathbf{W}(0))] - \mathbf{1}\left[ c\geq \Theta\left(\frac{n^5 \log^2 m}{\phi^2}\right)\right] \Omega \left( {\frac{\phi^2}{n^5 \log^2m}}c\right) + O \left(1\right),
\end{align*}
where we use the inequality $-\frac{a^2 t}{4}(2-2\frac{b}{a\sqrt{t}})^2 \leq -\frac{a^2 t}{4}\mathbf{1}[t\geq \frac{4b^2}{a^2}] $ at the last step. According to Allen-Zhu et al~\cite{allen2019convergence}, $\log [L( \mathbf{W}(0))] \leq O(n\log^2 m)$ with probability at least $1-e^{-\log^2 m}$, and using our choice $R \geq \Omega \left(\frac{ n^5\log^2m}{\phi^2} \log \frac{n\log^2 m}{\epsilon}\right)$ we have the following bound:
\begin{align*}
    \log [L(\mathbf{W}(T))]   
     & \leq O(n\log^2 m)  -   \Omega \left( \log \frac{n\log^2 m}{\epsilon}\right) \leq \log \epsilon,
\end{align*}
so we conclude that $L(\mathbf{W}(T)) \leq \epsilon$, or equavilently, $L(\mathbf{W}(T)) \leq  {n\log^2 m}\cdot e^{-R/R_0}$, where $R_0 = \frac{ n^5\log^2m}{\phi^2}$.

\vspace{-0.25cm}

\end{document}